\documentclass[lettersize,journal]{IEEEtran}

\usepackage{amsmath,amsfonts}
\usepackage{algorithm}
\usepackage{array}
\usepackage[caption=false,font=normalsize,labelfont=sf,textfont=sf]{subfig}
\usepackage{textcomp}
\usepackage{stfloats}
\usepackage{url}
\usepackage{verbatim}
\usepackage{graphicx}
\usepackage{cite}
\usepackage{multirow}
\usepackage{adjustbox}
\usepackage{booktabs}
\usepackage{makecell}
\usepackage{mathtools}

\usepackage{amsmath}
\usepackage{amssymb}
\usepackage{bm}
\usepackage{algorithm}
\usepackage{algpseudocode}

\algrenewcommand\algorithmicrequire{\textbf{Require:}}
\algrenewcommand\algorithmicensure{\textbf{Ensure:}}
\algrenewcommand\algorithmicreturn{\textbf{return}}

\usepackage[table]{xcolor}
\newcommand{\venue}[1]{%
    {\scriptsize\textcolor{gray}{ [#1]}}%
}

\begin{document}

\title{Calibrating Multimodal Test-Time Adaptation under Reliability and Availability Modality Shifts}
\author{
Jiangmeng Li, Jinglin Xu, Yi Li, Yumeng Liu, Jiahuan Zhou, \\
Fanjiang Xu, Hui Xiong, \IEEEmembership{Fellow,~IEEE}, Gang Hua, \IEEEmembership{Fellow,~IEEE}

\thanks{
Jiangmeng Li, Jinglin Xu, Yi Li, Yumeng Liu, and Fanjiang Xu are with the National Key Laboratory of Space Integrated Information System, Institute of Software, Chinese Academy of Sciences, and University of Chinese Academy of Sciences, Beijing, China; E-mail: \{jiangmeng2019, yumeng, fanjiang\}@iscas.ac.cn; \{xujinglin23, liyi212\}@mails.ucas.ac.cn.

Jiahuan Zhou is with the Wangxuan Institute of Computer Technology, Peking University, Beijing, China. E-mail: jiahuanzhou@pku.edu.cn.

Hui Xiong is with Thrust of Artificial Intelligence, the Hong Kong University of Science and Technology (Guangzhou), Guangzhou, China. He is also with Department of Computer Science \& Engineering, the Hong Kong University of Science and Technology, Hong Kong SAR, China. E-mail: xionghui@ust.hk.

Gang Hua is with Amazon Alexa AI, Bellevue, WA 98004 USA. E-mail: ganghua@gmail.com.

(Jiangmeng Li and Jinglin Xu contributed equally to this work.) 
(Corresponding author: Yi Li, Jiahuan Zhou)
}
}

% \markboth{IEEE TRANSACTIONS ON PATTERN ANALYSIS AND MACHINE INTELLIGENCE}
% {}
\markboth{Preprint.}
{}

\maketitle

\begin{abstract}
Multimodal test-time adaptation (TTA) enables pretrained multimodal models to accommodate distribution shifts during inference using only unlabeled test data, without recourse to the original training data.
Compared with unimodal adaptation, multimodal adaptation is further complicated by the coexistence and interaction of multiple representation perspectives.
This complexity is particularly pronounced under modality reliability shifts, which preserve modality availability but degrade information quality, and modality availability shifts, which render the corresponding modality entirely inaccessible.
Category-conditional Gaussian modeling offers a natural starting point for handling such distributional variations, having proven effective in unimodal test-time adaptation for characterizing representation distributions, yet its extension to multimodal settings requires a more elaborate formulation.
To this end, we propose Adaptive Probabilistic Gaussian Calibration (AdaPGC), which extends category-conditional Gaussian modeling across modality-specific and multimodal fusion representations for probabilistic calibration under multimodal distribution shifts.
AdaPGC continuously updates category-conditional Gaussian models over modality-specific and fusion representations, using the resulting probabilistic predictions to calibrate source model predictions and guide reliability based rectification, while exploiting their joint distributions to recover unavailable fusion representations.
Extensive experiments under diverse modality corruption and missingness conditions demonstrate that AdaPGC effectively improves robustness under both modality reliability and availability shifts.
Our source code will be released at https://github.com/XuJinglinn/AdaPGC-plus.
\end{abstract}

\begin{IEEEkeywords}
Multimodal Learning,
Test-Time Adaptation, 
Category-Conditional Gaussian Model.
\end{IEEEkeywords}

\section{Introduction}
\label{sec:intro}

\begin{figure*}[t]
\centering
\includegraphics[width=0.89\linewidth]{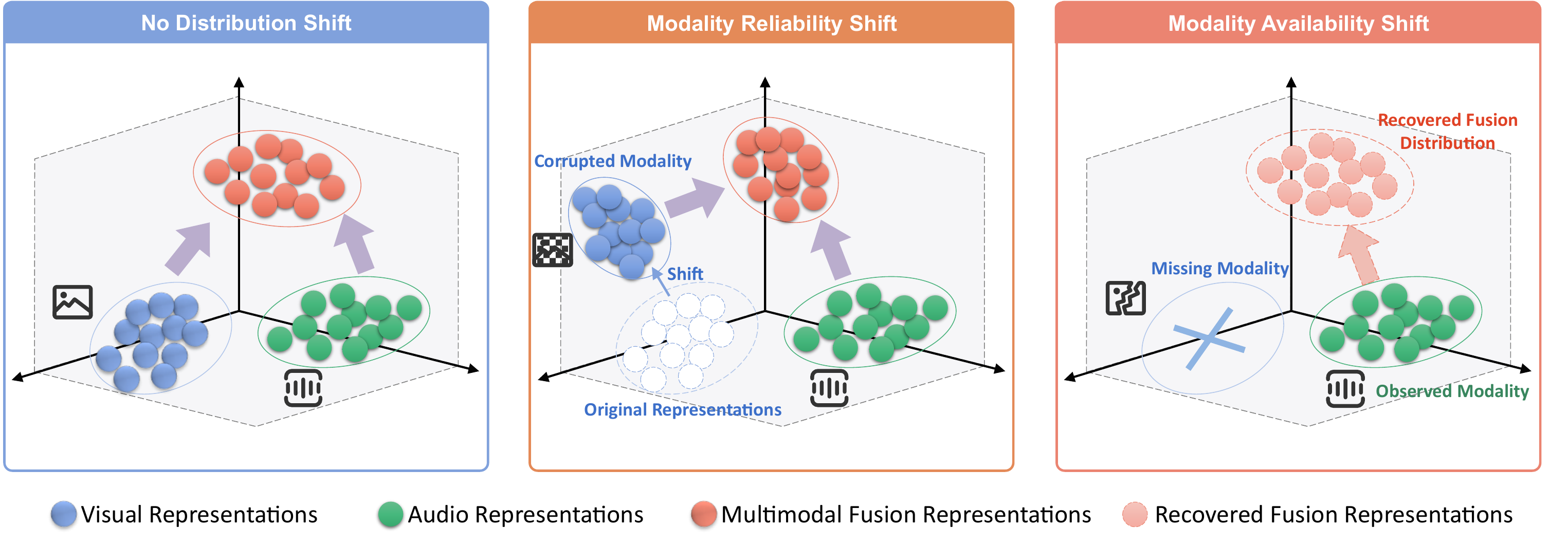}
\caption{
Illustration of modality reliability and availability shifts in multimodal TTA.
Under reliability shifts, corrupted modalities distort the corresponding modality-specific representations and consequently affect the multimodal fusion representation. 
Under availability shifts, missing modalities make the corresponding modality-specific representations unavailable and prevent the fusion representation from being directly constructed, motivating its recovery from the remaining modality.
}
\label{fig:motivation}
\end{figure*}

% Background: from multimodal learning to multimodal TTA 
\IEEEPARstart{M}{ultimodal} models integrate complementary information from heterogeneous data sources~\cite{xu2023multimodal,bruce2022mmnet,zhu2024vision+} and have demonstrated considerable promise in tasks such as action recognition~\cite{kazakos2019epic,nagrani2021attention,das2020mmhar,liu2020benchmark,sun2022human}, sentiment analysis~\cite{DBLP:conf/emnlp/ZadehCPCM17,DBLP:conf/acl/TsaiBLKMS19,yu2022unified,yu2021learning,li2025tracing}, and other complex perception tasks~\cite{anderson2018bottom,bai2022transfusion,xin2024mmap,zhang2024tamm,kamath2021mdetr,ding2022vlt,zhang2024vision}.
Conventional multimodal models are often vulnerable to distribution shifts, wherein discrepancies between the source and target domains can precipitate pronounced performance degradation~\cite{jaritz2022cross,chen2023benchmarking,dong2026advances}.
In response to this limitation, multimodal test-time adaptation (TTA) \cite{shin2022mm,DBLP:conf/iclr/Yang0Z0024,guo2025smoothing} has emerged as a viable paradigm, which dynamically adapts multimodal models using only unlabeled target data, thus facilitating real-time robustness enhancement under distribution shifts.

% Inspiration: 
% Category-conditional Gaussian modeling has been proven effective in unimodal TTA 
% but has challenges in extending to multimodal TTA: 
% (1)  multiple representations not independent but has specific information
% (2)  different and more complex distribution shift
Recent unimodal TTA methods~\cite{cui2025bayestta,han2026dota,dai2025free} model category-conditional distributions of test-time representations through Gaussian discriminant analysis (GDA)~\cite{hastie2009elements,wang2024hard} and have demonstrated consistent improvements under distribution shifts.
However, extending such distribution modeling to multimodal TTA is considerably more challenging.
Compared with unimodal models, multimodal models derive multiple modality-specific representations together with a multimodal fusion representation~\cite{li2021align,singh2022flava,dou2022empirical}, making it insufficient to treat the distribution modeled in each interrelated but distinct perspective merely as an independent basis for prediction.
Crucially, the coexistence of multiple representation perspectives renders test-time distribution shifts substantially more intricate in multimodal settings~\cite{li2026brimpr}.

% Question: 2 types of multimodal distribution shift
% modality reliability shifts & modality availability shifts
Specifically, as illustrated in Figure~\ref{fig:motivation}, test-time distribution shifts in multimodal settings are further complicated by \textit{modality reliability shifts} and \textit{modality availability shifts}.
Under \textit{modality reliability shifts}, the affected modality remains fully observable, while its information quality deteriorates due to noise, blur, environmental interference, or other corruptions.
Such degradation can distort the corresponding modality-specific representation, subsequently perturbing the multimodal fusion representation and ultimately compromising prediction performance~\cite{gao2024embracing}.
\textit{Modality availability shifts} constitute a more disruptive form of test-time variation, occurring when sensor failures, transmission interruptions, or privacy restrictions render the corresponding modality entirely unavailable for certain test instances~\cite{ma2022multimodal,wang2023distribution,ramazanova2025test,reza2024robust}.
In this case, not only is the representation of the missing modality inaccessible, but the corresponding multimodal fusion representation can no longer be directly constructed, resulting in a more severe loss of complementary information.
Addressing these two forms of distribution shift therefore requires a more nuanced exploitation of the information available across different representation perspectives.

% core idea
To address these challenges, we exploit the distributional information revealed by different representation perspectives in the target stream to calibrate source-model predictions under varying modality distribution shifts. 
More concretely, modality-specific and multimodal fusion representations characterize category structure from different information sources, and their corresponding distributional predictions therefore vary with changes in modality reliability~\cite{cao2023multi}.
When the reliability of a modality degrades, the resulting probabilistic predictions from different perspectives become relatively inconsistent. 
Such discrepancies can serve as an indicator of which modality is more strongly affected and thereby support targeted prediction calibration. 
Meanwhile, complete-modality test instances naturally provide simultaneously observed modality-specific and fusion representations, forming online paired observations in the target stream. 
These pairs provide a statistical basis for recovering the unavailable fusion representation from the observed modality. 
Such recovery can partially compensate for the multimodal predictive information lost when direct fusion becomes impossible due to modality missingness~\cite{li2024correlation,chen2024probabilistic}.

% method
Building on these considerations, we propose \textit{\textbf{Ada}ptive \textbf{P}robabilistic \textbf{G}aussian \textbf{C}alibration} (AdaPGC), 
\footnote{This work extends our preliminary conference paper~\cite{xu2026cvpr} in the following aspects: 
(1) we broaden the problem scope from modality reliability shifts under fully observed inputs to multimodal TTA under both reliability and availability variations;
(2) we augment the original probabilistic calibration framework with category-conditional joint distribution modeling, enabling cross-perspective statistical dependencies to support fusion representation inference under modality missingness; and
(3) we considerably expand the empirical evaluation to encompass a wider spectrum of modality corruption and missingness scenarios.}
a probabilistic approach for multimodal TTA under varying modality reliability and availability.
AdaPGC builds category-conditional Gaussian models for different representation perspectives and continuously updates them using statistics from the target stream. 
These models yield probabilistic predictions that provide distribution-aware calibration signals for the source model under different forms of modality shift.
Under modality reliability shifts, probabilistic predictions from different perspectives are compared to assess relative modality reliability, based on which we employ a contrastive rectification approach to correct the representation of the less reliable modality.
Under modality availability shifts, AdaPGC further extends the category-conditional Gaussian modeling to the joint distribution of modality-specific and multimodal fusion representations, with the model parameters learned from modality-complete test instances.
This extension yields a closed-form conditional estimate of the unavailable fusion representation from the observed modality. This estimate is then incorporated into the probabilistic calibration of source model predictions.
Overall, our contributions can be summarized as follows:
\begin{itemize}
    \item We study multimodal TTA under both modality reliability and availability shifts, covering scenarios where modalities remain observable but vary in reliability, as well as scenarios where part of the multimodal input becomes unavailable. 
    This setting allows us to systematically investigate robustness to both corrupted and incomplete multimodal observations.
    \item We propose AdaPGC, a probabilistic approach for multimodal TTA under both modality reliability and availability shifts. 
    AdaPGC models both category-conditional distributions and the dependencies between modality-specific and fusion representations, enabling a unified probabilistic treatment of fully observed, corrupted, and partially observed multimodal inputs.
    \item We conduct extensive experiments under diverse modality corruption and missingness conditions, demonstrating the effectiveness and robustness of AdaPGC under both modality reliability and availability shifts.
    
\end{itemize}

\section{Related work}
\label{sec:related-work}
\subsection{Multimodal Test-Time Adaptation }

Test-time adaptation (TTA) adapts a pretrained model to distribution shifts using only unlabeled target data. 
Representative unimodal approaches include entropy minimization with TENT~\cite{wang2021tent}, informative sample selection with EATA~\cite{pmlr-v162-niu22a} and SAR~\cite{niu2023towards}, and test-time classifier adjustment with T3A~\cite{iwasawa2021test}.
These studies establish the basic adaptation mechanisms in unimodal settings, while multimodal TTA further considers the complementary information and interactions among different modalities.

Building on this setting, existing multimodal TTA methods exploit cross-modal information through diverse adaptation strategies.
MM-TTA~\cite{shin2022mm} generates complementary pseudo-labels across modalities to facilitate source-free adaptation, although its effectiveness depends on the quality of the generated pseudo-labels.
READ~\cite{DBLP:conf/iclr/Yang0Z0024} introduces a more robust approach by enhancing the impact of confident predictions and reducing the influence of noisy predictions. 
Meanwhile, ABPEM~\cite{zhao2025attention} improves multimodal TTA by mitigating modality misalignment under distribution shift through attention bootstrapping and principal entropy minimization.
More recently, SuMi~\cite{guo2025smoothing} employs smoothing and cross-modal information sharing to improve robustness under mixed shifts, including modality corruption and missingness.
TSA~\cite{chen2025testtime}, a recent method, further explores the problem of unimodal distribution shift through a selective adaptation framework that handles shifts in the affected modality.
BriMPR~\cite{li2026brimpr} addresses the coupled effects of unimodal feature shifts and cross-modal semantic misalignment through progressive re-alignment.

Most existing methods are predicated on full modality observability and thus do not accommodate modality missing scenarios.
In contrast, AdaPGC jointly models marginal and cross-representation distributions, enabling unified adaptation to modality reliability and availability shifts.

\subsection{Incomplete Multimodal Learning}

Incomplete multimodal learning seeks to preserve predictive robustness under partially observed inputs, where variations in modality availability can otherwise result in substantial performance degradation.
Existing studies have developed various mechanisms for learning with incomplete multimodal inputs. 
MAP~\cite{lee2023multimodal} introduces modality-missing-aware prompts to equip pretrained multimodal transformers for diverse missing-modality conditions.
DCP~\cite{hu2024deep} further models the correlations between prompts and input features, as well as those across different prompt layers, to better exploit cross-modal complementary information.
More recently, MoRA~\cite{zhao2026mora} introduces modality-common and modality-specific low-rank parameters to facilitate bidirectional cross-modal knowledge transfer while preserving modality-specific flexibility.

A limited number of studies have investigated missing modalities under the more restrictive TTA setting.
MiDl~\cite{ramazanova2025test} combines mutual information minimization with self-distillation to adapt multimodal models using incomplete test inputs. 
SuMi~\cite{guo2025smoothing} handles missing modalities by smoothing distribution shifts between complete and incomplete observations.

Despite this progress, existing methods primarily enhance predictions from the available modalities, without explicitly recovering the inaccessible fusion representation. 
AdaPGC addresses this limitation by modeling the category-conditional distributions of modality-specific and fusion representations, enabling the missing fusion representation to be inferred from the observed modality.

\subsection{Distribution Modeling for TTA}
Distribution modeling provides a probabilistic basis for test-time adaptation by estimating the category structure of unlabeled target representations. 
In particular, category-conditional Gaussian modeling enables posterior inference through Gaussian discriminant analysis (GDA), using feature statistics accumulated during inference.
DOTA~\cite{han2026dota} continuously estimates category-conditional distributions and derives posterior probabilities via Bayesian inference, mitigating the forgetting associated with limited sample caches. 
FreeTTA~\cite{dai2025free} employs online expectation-maximization to estimate target distributions, incorporating zero-shot predictions as prior information.
BayesTTA~\cite{cui2025bayestta} further considers temporally evolving distributions, combining incremental Gaussian estimation with adaptive covariance selection to calibrate predictions and guide representation adaptation. 
ADAPT~\cite{zhangbackpropagation} incorporates CLIP priors and a historical knowledge bank to correct potential likelihood biases, refining GDA-based inference without backpropagation.
These studies demonstrate the value of target feature statistics for informing test-time predictions.

These approaches primarily model distributions in a single visual representation space. 
AdaPGC extends category-conditional Gaussian modeling to modality-specific and fusion representations, using their marginal distributions for probabilistic calibration and their joint distributions for fusion representation recovery.

\section{Preliminary}

\subsection{Gaussian Discriminant Analysis}
Gaussian Discriminant Analysis (GDA) is a classical generative classification 
model that predicts category labels based on category-conditional feature 
distributions~\cite{hastie2009elements,murphy2022probabilistic}. 
Consider a labeled sample set
$\mathcal{D}=\{(\boldsymbol{z}_i,y_i)\}_{i=1}^{N}$, where
$\boldsymbol{z}_i\in\mathbb{R}^{d}$ is a $d$-dimensional feature vector and
$y_i\in\{1,2,\ldots,C\}$ is its category label, with $N$ and $C$ denoting the
number of samples and categories, respectively. GDA assumes that the feature
distribution conditioned on category $c$ is Gaussian:
\begin{equation}
\begin{aligned}
p(\boldsymbol{z}_i\mid c)
&=\mathcal{N}(\boldsymbol{z}_i;\boldsymbol{\mu}_c,\boldsymbol{\Sigma}_c)\\
&=\frac{
\exp\left(
-\frac{1}{2}
(\boldsymbol{z}_i-\boldsymbol{\mu}_c)^\top
\boldsymbol{\Sigma}_c^{-1}
(\boldsymbol{z}_i-\boldsymbol{\mu}_c)
\right)}
{\sqrt{(2\pi)^d\lvert\boldsymbol{\Sigma}_c\rvert}},
\end{aligned}
\end{equation}
where $\boldsymbol{\mu}_c\in\mathbb{R}^{d}$ and
$\boldsymbol{\Sigma}_c\in\mathbb{R}^{d\times d}$ are the mean vector and
covariance matrix of category $c$, respectively.

According to Bayes' theorem, the posterior probability that
$\boldsymbol{z}_i$ belongs to category $c$ is
\begin{equation}
p(c\mid\boldsymbol{z}_i)
=
\frac{
\pi_c\,p(\boldsymbol{z}_i\mid c)}
{\sum_{c'=1}^{C}\pi_{c'}\,p(\boldsymbol{z}_i\mid c')},
\end{equation}
where $\pi_c=p(c)$ denotes the prior probability of category $c$.

Taking the logarithm of the posterior probability above and omitting terms shared across all categories yields the following log-posterior for category $c$:

\begin{equation}
\label{eq:gda-score}
\begin{aligned}
g_c(\boldsymbol{z}_i) ={}&
-\frac{1}{2}
(\boldsymbol{z}_i-\boldsymbol{\mu}_c)^\top
\boldsymbol{\Sigma}_c^{-1}
(\boldsymbol{z}_i-\boldsymbol{\mu}_c)\\
&-\frac{1}{2}\log\lvert\boldsymbol{\Sigma}_c\rvert
+\log\pi_c.
\end{aligned}
\end{equation}
The posterior probabilities can be recovered by applying the softmax
function to the log-posteriors.

\subsection{Conditional Gaussian Estimation}

Let $\boldsymbol{z}_X$ and $\boldsymbol{z}_Y$ denote two statistically correlated representation vectors. 
Suppose their statistical dependence is characterized by the following joint Gaussian distribution:
\begin{equation}
\begin{bmatrix}
\boldsymbol z_X\\
\boldsymbol z_Y
\end{bmatrix}
\sim
\mathcal N
\left(
\begin{bmatrix}
\boldsymbol\mu_X\\
\boldsymbol\mu_Y
\end{bmatrix},
\begin{bmatrix}
\boldsymbol\Sigma_X & \boldsymbol\Sigma_{XY}\\
\boldsymbol\Sigma_{YX} & \boldsymbol\Sigma_Y
\end{bmatrix}
\right),
\end{equation}
where $\boldsymbol{\mu}_X$ and $\boldsymbol{\mu}_Y$ are the corresponding mean vectors,
$\boldsymbol{\Sigma}_X$ and $\boldsymbol{\Sigma}_Y$ denote their covariance matrices, and
$\boldsymbol{\Sigma}_{YX}
=
\boldsymbol{\Sigma}_{XY}^{\top}$
denotes the cross-covariance between the two representations.

A fundamental property of the joint Gaussian distribution is that the conditional distribution of one representation given the other remains Gaussian~\cite{williams2006gaussian}. Specifically,
\begin{equation}
\boldsymbol z_Y \mid \boldsymbol z_X
\sim
\mathcal N
\left(
\boldsymbol\mu_{Y\mid X},
\boldsymbol\Sigma_{Y\mid X}
\right),
\end{equation}
with conditional mean and covariance
\begin{equation}
\boldsymbol\mu_{Y\mid X}
=
\boldsymbol\mu_Y
+
\boldsymbol\Sigma_{YX}
\boldsymbol\Sigma_X^{-1}
\left(
\boldsymbol z_X-\boldsymbol\mu_X
\right),
\end{equation}
\begin{equation}
\boldsymbol\Sigma_{Y\mid X}
=
\boldsymbol\Sigma_Y
-
\boldsymbol\Sigma_{YX}
\boldsymbol\Sigma_X^{-1}
\boldsymbol\Sigma_{XY}.
\end{equation}

The conditional mean gives the expected representation of $\boldsymbol z_Y$ for an observed $\boldsymbol z_X$. 
It can therefore be used as a point estimate of the unobserved representation~\cite{ghahramani1993supervised,murphy2022probabilistic}. 
We therefore define the conditional Gaussian estimation function as
\begin{equation}
\label{eq:gaussian-estimation}
\begin{aligned}
\mathcal E_{X\to Y}(\boldsymbol z_X)
&=
\mathbb E
\left[
\boldsymbol z_Y \mid \boldsymbol z_X
\right]\\
&=
\boldsymbol\mu_Y
+
\boldsymbol\Sigma_{YX}
\boldsymbol\Sigma_X^{-1}
\left(
\boldsymbol z_X-\boldsymbol\mu_X
\right).
\end{aligned}
\end{equation}
The unobserved representation can thus be estimated as
$\hat{\boldsymbol z}_{Y\mid X} = \mathcal E_{X\to Y}(\boldsymbol z_X)$,
using the statistical dependence captured by the cross-covariance.

\section{Method}
\begin{figure*}[t]
\centering
\includegraphics[width=\linewidth]{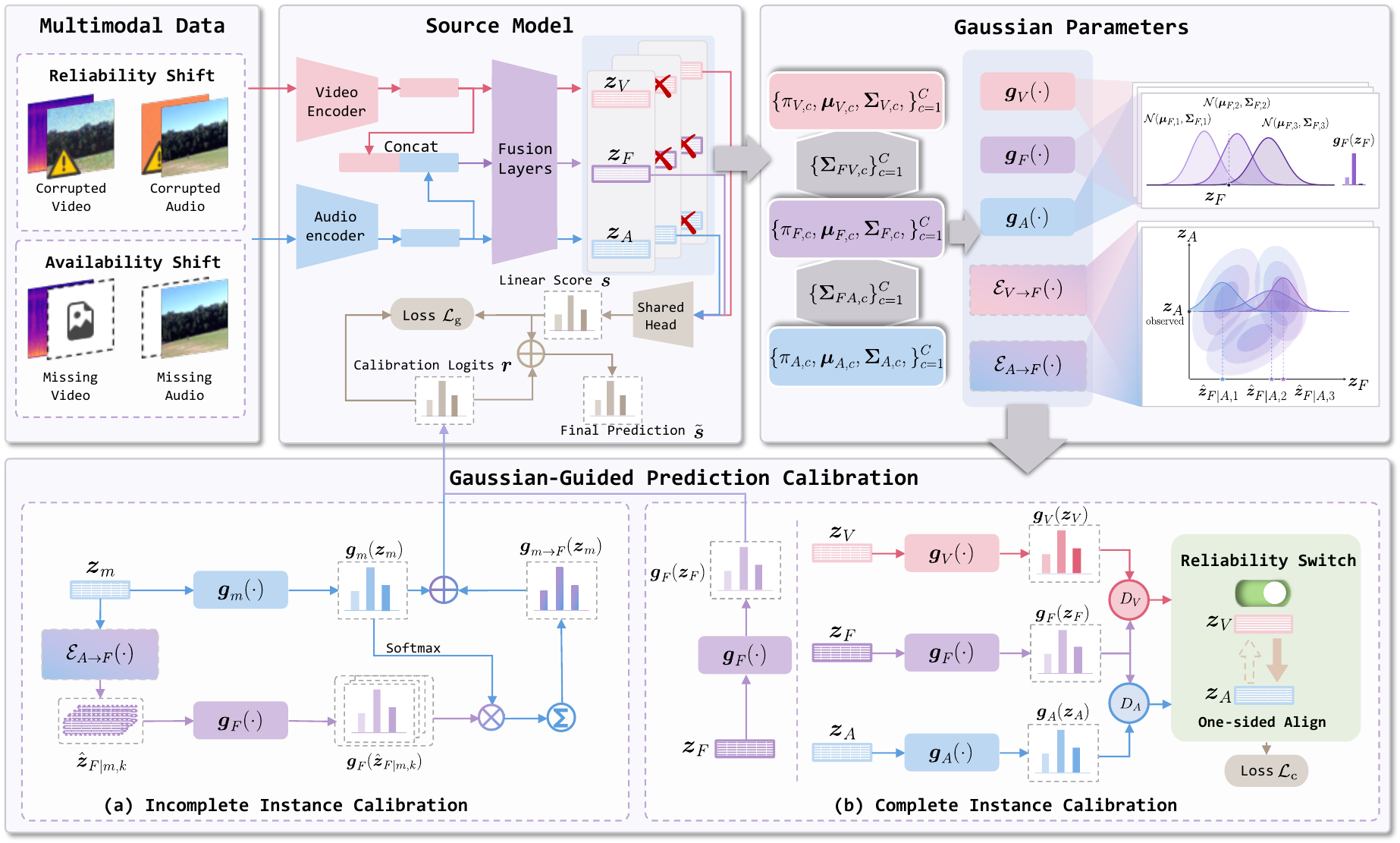}
\caption{
Overview of the proposed multimodal test-time adaptation framework.
The input consists of paired video and audio streams affected by reliability shifts (modality corruption) or availability shifts (missing video/audio). 
For complete instances, the pretrained source model feeds the fused representation \(z_F\) into the shared classifier to obtain source logits \(s\); for incomplete instances, it instead uses the representation of the available modality, \(z_V\) or \(z_A\).
The category-conditional Gaussian means and priors are initialized from the pretrained classifier, while the covariance and cross-covariance matrices are initialized as identity matrices. During adaptation, these parameters are updated using target-stream statistics to define the discriminants $g_V$, $g_A$, and $g_F$ and conditional estimators $\mathcal{E}_{V\rightarrow F}$ and $\mathcal{E}_{A\rightarrow F}$.
For incomplete instances, the available modality representation is used to estimate category-conditional fused representations, whose Gaussian predictions are combined with the unimodal predictions to obtain calibration logits \(r\).
For complete instances, the fused Gaussian discriminant produces the calibration logits \(r=g_F(z_F)\), while discrepancies between unimodal and fused Gaussian predictions guide reliability-aware one-sided alignment.
}
\label{fig:arch}
\end{figure*}

This section presents the proposed AdaPGC framework. 
We first formulate multimodal test-time adaptation under modality distribution shifts, covering modality degradation and modality missingness. 
We then introduce online category-conditional Gaussian modeling for modality-specific and fusion representations, which provides the probabilistic foundation for subsequent adaptation. 
Based on these Gaussian distributions, we develop Gaussian probabilistic calibration to refine source-model predictions, where the joint distributions between modality-specific and fusion representations are further exploited to recover unavailable fusion representations under modality missingness. 
Finally, for samples affected by single-modality degradation, we introduce adaptive contrastive rectification to identify the relatively degraded modality and selectively rectify its representation while preserving the reliable modality.
The overall framework is illustrated in Figure~\ref{fig:arch}.

\subsection{Problem Formulation}
Without loss of generality, we consider a multimodal classification problem with two modalities, denoted by $m_1$ and $m_2$. 
At test time, the unlabeled target data arrive sequentially as a stream of mini-batches,
$\mathcal{D}_{\mathrm{target}}=\{\mathcal{D}^{t}\}_{t=1}^{T}$,
where $t$ indexes the incoming batch, $T$ denotes the total number of test batches, and each batch contains $\mathcal{B}$ samples.
Neither the source-domain data nor the target labels are accessible, and the model is updated sequentially without resetting its parameters between batches.
Depending on modality availability, each batch is partitioned into three disjoint subsets:
\begin{equation}
\begin{aligned}
\mathcal{D}^{t}
  &= \mathcal{D}^{t}_{\mathrm{full}}
     \mathbin{\cup}
     \mathcal{D}^{t}_{m_1}
     \mathbin{\cup}
     \mathcal{D}^{t}_{m_2}, \\[-1pt]
\left|\mathcal{D}^{t}_{\mathrm{full}}\right|
  &+
  \left|\mathcal{D}^{t}_{m_1}\right|
  +
  \left|\mathcal{D}^{t}_{m_2}\right|
  = \mathcal{B}.
\end{aligned}
\label{eq:batch-partition}
\end{equation}
Here, $\mathcal{D}^{t}_{\mathrm{full}}$ contains samples for which both modalities $m_1$ and $m_2$ are available, i.e., paired observations $(\boldsymbol{x}^{t}_{m_1,i},\boldsymbol{x}^{t}_{m_2,i})$. In contrast, $\mathcal{D}^{t}_{m_1}$ contains samples for which only modality $m_1$ is available, while $\mathcal{D}^{t}_{m_2}$ contains samples for which only modality $m_2$ is available.

Let
$\mathcal{M}_{\theta}
=
(\phi_1,\phi_2,\mathcal{F},s)$
denote the pretrained multimodal model, where $\phi_1$ and $\phi_2$ are the modality encoders, $\mathcal{F}$ denotes the fusion layers, and $s$ is the shared final linear classifier. 
The source model supports both unimodal and multimodal inference. 
% We take the representations produced by the fusion layers immediately before $s$ as features. 
For a modality-complete sample, the model produces
\begin{align}
\boldsymbol{z}_{m_1,i}^{t}
&=
\mathcal{F}\!\left(
\phi_1(\boldsymbol{x}_{m_1,i}^{t})
\right), \\
\boldsymbol{z}_{m_2,i}^{t}
&=
\mathcal{F}\!\left(
\phi_2(\boldsymbol{x}_{m_2,i}^{t})
\right), \\
\boldsymbol{z}_{F,i}^{t}
&=
\mathcal{F}\!\left(
\operatorname{Concat}\!\left[
\phi_1(\boldsymbol{x}_{m_1,i}^{t}),
\phi_2(\boldsymbol{x}_{m_2,i}^{t})
\right]
\right).
\end{align}
Here, $\boldsymbol{z}_{m_1,i}^{t}$ and
$\boldsymbol{z}_{m_2,i}^{t}$ are the unimodal representations, while
$\boldsymbol{z}_{F,i}^{t}$ is the fused representation.
For a modality-incomplete sample, only the unimodal representation corresponding to the available modality can be obtained. Specifically, a sample in $\mathcal{D}_{m_1}^{t}$ produces $\boldsymbol{z}_{m_1,i}^{t}$, while a sample in $\mathcal{D}_{m_2}^{t}$ produces $\boldsymbol{z}_{m_2,i}^{t}$. In either case, the fused representation $\boldsymbol{z}_{F,i}^{t}$ cannot be directly obtained because one modality is unavailable.

Since the three representations are mapped to category logits by the same linear classifier $s$, we regard them as three perspectives of the same sample in a shared representation space.
To describe the classification of different representations in a
unified manner, let
$m\in\{m_1,m_2,F\}$
index a representation perspective. For any available representation
$\boldsymbol{z}_{m,i}^{t}$, the logit of category $c$ is given by
\begin{equation}
s_c\!\left(\boldsymbol{z}_{m,i}^{t}\right)
=
\boldsymbol{w}_c^{\top}\boldsymbol{z}_{m,i}^{t}
+
b_c,
\qquad
c\in\{1,\ldots,C\},
\end{equation}
where $\boldsymbol{w}_c$ and $b_c$ are the classifier weight and bias
for category $c$, respectively.
Collecting the category-wise logits into
$\boldsymbol{s}(\boldsymbol{z})=[s_1(\boldsymbol{z}),\ldots,s_C(\boldsymbol{z})]^{\top}$, the source logits for each target sample are determined by its modality availability as
\begin{equation}
\boldsymbol{s}_{i}^{t}
=
\begin{cases}
\boldsymbol{s}\left(\boldsymbol{z}_{F,i}^{t}\right),
&
\boldsymbol{x}_{i}^{t}\in\mathcal{D}_\text{full}^{t},
\\[4pt]
\boldsymbol{s}\left(\boldsymbol{z}_{m_1,i}^{t}\right),
&
\boldsymbol{x}_{i}^{t}\in\mathcal{D}_{m_1}^{t},
\\[4pt]
\boldsymbol{s}\left(\boldsymbol{z}_{m_2,i}^{t}\right),
&
\boldsymbol{x}_{i}^{t}\in\mathcal{D}_{m_2}^{t}.
\end{cases}
\end{equation}
The corresponding prediction is
$\hat{y}_{i}^{t}=\arg\max_{c\in\{1,\ldots,C\}}\boldsymbol{s}_{c,i}^{t}$.
Thus, the objective is to continuously adapt the pretrained model to the incoming target stream while maintaining accurate predictions under diverse modality shifts, including sample-wise modality missingness.

\subsection{Test-Time Gaussian Modeling}

The source model provides discriminative predictions through learned decision boundaries, but such predictions alone do not explicitly capture the underlying feature distributions under target shifts.
To obtain distribution-aware evidence for test-time adaptation, we establish a category-conditional Gaussian modeling framework over target representations.

Specifically, for each perspective $m\in\{m_1,m_2,F\}$ and category $c$, we assume a category-conditional Gaussian distribution 
\begin{equation}
p(\boldsymbol{z}_{m,i}^t \mid y_{i} = c)
=
\mathcal{N}
\left(
\boldsymbol{z}_{m}^t;
\boldsymbol{\mu}_{m,c}^t,
\boldsymbol{\Sigma}_{m,c}^t
\right),
\label{eq:category_gaussian}
\end{equation}
together with the corresponding category prior $\pi_{m,c}^t=p(c)$, 
where $\boldsymbol{\mu}_{m,c}^t$, $\boldsymbol{\Sigma}_{m,c}^t$ and $\pi_{m,c}^t$ denote the category-specific mean vector, covariance matrix, and prior probability for perspective $m$, respectively, estimated from the target samples observed up to the $t$-th batch. 
Applying Bayes' rule and omitting terms shared across categories gives the  log-posterior score for each perspective:
\begin{equation}
\begin{aligned}
g_{m,c}^t(\boldsymbol{z}_{m,i}^t)
=&
-\frac{1}{2}
(\boldsymbol{z}_{m,i}^t-\boldsymbol{\mu}_{m,c}^t)^\top
(\boldsymbol{\Sigma}_{m,c}^t)^{-1}
(\boldsymbol{z}_{m,i}^t-\boldsymbol{\mu}_{m,c}^t)\\
&-\frac{1}{2}\log |\boldsymbol{\Sigma}_{m,c}^t|
+\log \pi_{m,c}^t.
\end{aligned}
\end{equation}
Collecting the scores over all categories gives $\boldsymbol{g}_{m}^t(\boldsymbol{z}_{m,i}^t)$, resulting in three perspective-specific probabilistic predictions for $m_1$, $m_2$, and $F$ after softmax.

Moreover, to model the statistical dependency between different representations, we further assume that their joint distribution conditioned on category $c$ follows a joint Gaussian distribution. 
For each pair consisting of a modality-specific representation $\boldsymbol{z}_{m,i}^t$, with $m\in\{m_1,m_2\}$, and the corresponding fusion representation $\boldsymbol{z}_{F,i}^t$ from a modality-complete sample, we specify their joint distribution as follows:
\begin{equation}
\begin{bmatrix}
\boldsymbol{z}_{m,i}^t\\
\boldsymbol{z}_{F,i}^t
\end{bmatrix}
\Bigg|c
\sim
\mathcal N
\left(
\begin{bmatrix}
\boldsymbol\mu_{m,c}^t\\
\boldsymbol\mu_{F,c}^t
\end{bmatrix},
\begin{bmatrix}
\boldsymbol\Sigma_{m,c}^t
&
{\boldsymbol\Sigma_{Fm,c}^t}^{\top}
\\
\boldsymbol\Sigma_{Fm,c}^t
&
\boldsymbol\Sigma_{F,c}^t
\end{bmatrix}
\right).
\label{eq:category_joint_gaussian}
\end{equation}
Among these parameters, the mean vectors \(\boldsymbol{\mu}_{m,c}^t\) and \(\boldsymbol{\mu}_{F,c}^t\), together with the covariance terms within each perspective, \(\boldsymbol{\Sigma}_{m,c}^t\) and \(\boldsymbol{\Sigma}_{F,c}^t\), are already obtained from the Gaussian modeling above.
Therefore, only the cross-covariance matrix \(\boldsymbol{\Sigma}_{Fm,c}^{t}\) needs to be additionally estimated from modality-complete samples, reflecting the statistical relationship between the modality-specific and fusion representations for each category.

Under this joint Gaussian assumption, we define the category-specific fusion estimation function as the conditional expectation of the fusion representation given the observed modality representation:
\begin{equation}
\label{eq:category_conditional_estimation}
\begin{aligned}
\hat{\boldsymbol z}_{F\mid m,i,c}^{t}
&=
\mathcal E_{m\to F,c}^{t}
\left(\boldsymbol z_{m,i}^{t}\right)\\
&=
\boldsymbol\mu_{F,c}^{t}
+
\boldsymbol\Sigma_{Fm,c}^{t}
\left(\boldsymbol\Sigma_{m,c}^{t}\right)^{-1}
\left(
\boldsymbol z_{m,i}^{t}-\boldsymbol\mu_{m,c}^{t}
\right).
\end{aligned}
\end{equation}
Here, $\hat{\boldsymbol z}_{F\mid m,i,c}^{t}$ is the conditional mean of the fusion representation given $\boldsymbol z_{m,i}^{t}$ and category $c$.

Since the category of a test instance is unknown, we apply the estimation function for each category to obtain a set of category-conditioned fusion estimates, denoted by
$\mathcal E_{m\to F}^{t}(\boldsymbol z_{m,i}^{t})
=\{\hat{\boldsymbol z}_{F\mid m,i,c}^{t}\}_{c=1}^{C}$.
         
\subsubsection{Gaussian Parameter Initialization}
A fundamental challenge in online Gaussian modeling is how to obtain informative priors for the category-conditional parameters before sufficient target samples have been observed. 
Since source data are unavailable during test-time adaptation, we instead exploit the category knowledge already encoded in the pretrained classifier to initialize these parameters.

For a category-conditional Gaussian distribution, the mean naturally determines the center of each category in the representation space. 
Meanwhile, pretrained classifier weights have been shown to encode category prototype information~\cite{iwasawa2021test}. 
This correspondence allows the pretrained classifier to provide an informative initialization for the category-conditional Gaussian parameters.
We further adopt an isotropic covariance initialization, avoiding the introduction of additional category-specific structure before target statistics become available.
For each $m\in\{m_1,m_2,F\}$ and category $c$, we initialize the Gaussian mean and covariance as
\begin{equation}
\boldsymbol{\mu}_{m,c}^{0}
=
\boldsymbol{w}_c
,
\qquad
\boldsymbol{\Sigma}_{m,c}^{0}
=
\boldsymbol{I}.
\label{eq:init_mu_cov}
\end{equation}

The category prior required by the Gaussian model can be naturally derived from the bias term of the pretrained classifier, which already reflects category-specific preference. 
Accordingly, we initialize the category prior by absorbing the classifier bias term into the category prior:
\begin{equation}
\log\pi_{m,c}^{0}
=
b_c
+
\frac{1}{2}
\|\boldsymbol{w}_c\|_2^2.
\label{eq:init_pi}
\end{equation}

With the above initialization, the initial log-posterior is equivalent to the original classifier score, i.e.,
$
g_{m,c}^{0}(\boldsymbol z)
=
s_c(\boldsymbol z).
$

For each modality $m\in\{m_1,m_2\}$, the category-conditional cross-covariance between its corresponding representation and the fusion representation is initialized as
\begin{equation}
\boldsymbol{\Sigma}_{Fm,c}^{0}
=
\boldsymbol I.
\end{equation}
Since the modality-specific and fusion representations are extracted from the same sample and lie in the same representation space, the identity matrix is adopted as the initial cross-covariance prior.

\subsubsection{Incremental Gaussian Statistics Update}

In multimodal TTA, target samples arrive sequentially, and their distribution often differs significantly from the source-domain distribution. 
Consequently, the initial parameters $\pi_{m,c}^0, \boldsymbol{\mu}_{m,c}^0, \boldsymbol{\Sigma}_{m,c}^0$ and $\boldsymbol{\Sigma}_{Fm,c}^0$ obtained from the pretrained source model may no longer be reliable.
To mitigate this mismatch, we progressively update these parameters with each incoming batch of target data, allowing the model to adapt to the evolving target distribution.

Since category labels are unavailable during test-time adaptation, we estimate the category assignment of each sample through soft responsibilities derived from the pretrained model predictions.
Given the source model logits $\boldsymbol{s}_{i}^{t}$, the soft responsibility of sample $i$ is defined as 
$
\boldsymbol{\gamma}_{i}^{t}
=
\operatorname{Softmax}
(
\boldsymbol{s}_{i}^{t}
).
$

For each category $c$ and perspective $m\in{m_1,m_2,F}$, we maintain sufficient statistics weighted by the corresponding category responsibilities, including the effective sample count $N_{m,c}^{t}$, first-order statistic $\boldsymbol{S}_{m,c}^{t}\in\mathbb{R}^{d}$, and second-order statistic $\boldsymbol{Q}_{m,c}^{t}\in\mathbb{R}^{d\times d}$. 
For each unimodal perspective $m\in{m_1,m_2}$, we additionally maintain the cross statistic $\boldsymbol{Q}_{Fm,c}^{t}\in\mathbb{R}^{d\times d}$. 
These statistics are computed as
\begin{equation*}
\begin{aligned}
N_{m,c}^{t}
&=
\sum_{i\in\mathcal{D}^{1:t}}
\gamma_{i,c}^{t},
\\
\boldsymbol{S}_{m,c}^{t}
&=
\sum_{i\in\mathcal{D}^{1:t}}
\gamma_{i,c}^{t}
\boldsymbol{z}_{m,i}^{t},
\\
\boldsymbol{Q}_{m,c}^{t}
&=
\sum_{i\in\mathcal{D}^{1:t}}
\gamma_{i,c}^{t}
\boldsymbol{z}_{m,i}^{t}
(\boldsymbol{z}_{m,i}^{t})^{\top},
\\
\boldsymbol{Q}_{Fm,c}^{t}
&=
\sum_{i\in\mathcal{D}^{1:t}}
\gamma_{i,c}^{t}
\boldsymbol{z}_{F,i}^{t}
(\boldsymbol{z}_{m,i}^{t})^{\top},
\end{aligned}
\end{equation*}
where $\mathcal{D}^{1:t}=\bigcup_{r=1}^{t}\mathcal{D}^{r}$ denotes all target samples observed up to batch $t$.
To enable streaming adaptation, the statistics are updated incrementally with the newly arrived batch:
$
\begin{aligned}
N_{m,c}^{t} &= N_{m,c}^{t-1} + \Delta N_{m,c}^{t}, &
\Delta N_{m,c}^{t} &= \sum_{i\in\mathcal{D}^{t}} \gamma_{ic}^{t}, \\
\mathbf{S}_{m,c}^{t} &= \mathbf{S}_{m,c}^{t-1} + \Delta \mathbf{S}_{m,c}^{t}, &
\Delta \mathbf{S}_{m,c}^{t} &= \sum_{i\in\mathcal{D}^{t}} \gamma_{ic}^{t}\,\boldsymbol{z}_{m,i}^t, \\
\mathbf{Q}_{m,c}^{t} &= \mathbf{Q}_{m,c}^{t-1} + \Delta \mathbf{Q}_{m,c}^{t}, &
\Delta \mathbf{Q}_{m,c}^{t} &= \sum_{i\in\mathcal{D}^{t}} \gamma_{ic}^{t}\,\boldsymbol{z}_{m,i}^t \boldsymbol{z}_{m,i}^{t\top}, \\
\mathbf{Q}_{Fm,c}^{t} &= \mathbf{Q}_{Fm,c}^{t-1} + \Delta \mathbf{Q}_{Fm,c}^{t}, &
\Delta \mathbf{Q}_{Fm,c}^{t} &= \sum_{i\in\mathcal{D}^{t}} \gamma_{ic}^{t}\,\boldsymbol{z}_{F,i}^t \boldsymbol{z}_{m,i}^{t\top}. 
\end{aligned}
$

Given these accumulated sufficient statistics, the maximum likelihood estimates of the category prior, mean, covariance, and cross-covariance have the following closed-form expressions:
\begin{equation}
\begin{aligned}
\hat{\pi}_{m,c}^t & = \frac{N_{m,c}^t}{\sum_{c^{'}} N_{m,c^{'}}^t},\\
\hat{\boldsymbol{\mu}}_{m,c}^t & =\, \frac{\mathbf{S}_{m,c}^t}{N_{m,c}^t},\\
\hat{\boldsymbol{\Sigma}}_{m,c}^t & =\, \frac{\mathbf{Q}_{m,c}^t}{N_{m,c}^t}\;-\;\hat{\boldsymbol{\mu}}_{m,c}^t\hat{\boldsymbol{\mu}}_{m,c}^{t\top},\\
\hat{\boldsymbol{\Sigma}}_{Fm,c}^t & =\, \frac{\mathbf{Q}_{Fm,c}^t}{N_{m,c}^t}\;-\;\hat{\boldsymbol{\mu}}_{F,c}^t\hat{\boldsymbol{\mu}}_{m,c}^{t\top}.\\
\end{aligned}
\end{equation}

For stable online parameter estimation, we adopt exponential moving average (EMA) updates with rate $\alpha$:
\begin{equation}
\begin{alignedat}{5}
&\pi_{m,c}^{t}
&&=\, \alpha\,\pi_{m,c}^{t-1}
&&+\; (1-\alpha)\,\hat{\pi}_{m,c}^{t},
\\
&\boldsymbol{\mu}_{m,c}^{t}
&&=\, \alpha\,\boldsymbol{\mu}_{m,c}^{t-1}
&&+\; (1-\alpha)\,\hat{\boldsymbol{\mu}}_{m,c}^{t},
\\
&\boldsymbol{\Sigma}_{m,c}^{t}
&&=\, \alpha\,\boldsymbol{\Sigma}_{m,c}^{t-1}
&&+\; (1-\alpha)\,\hat{\boldsymbol{\Sigma}}_{m,c}^{t},
\\
&\boldsymbol{\Sigma}_{Fm,c}^{t}
&&=\, \alpha\,\boldsymbol{\Sigma}_{Fm,c}^{t-1}
&&+\; (1-\alpha)\,\hat{\boldsymbol{\Sigma}}_{Fm,c}^{t}.
\end{alignedat}
\end{equation}

Overall, this online estimation enables the Gaussian parameters of modality-specific and fusion representations, together with the cross-covariance between them, to adapt to evolving test distributions without retaining historical samples.

\begin{algorithm*}[t]
\caption{Streaming Test-Time Adaptation of AdaPGC}
\label{alg:adapgc}
\begin{algorithmic}[1]
\Require Target stream
$\mathcal{D}_{\mathrm{target}}
=\{\mathcal{D}^{t}\}_{t=1}^{T}$;
pretrained model $\mathcal{M}_{\theta^{0}}$;
loss weights $w_{\text{g}},w_{\text{c}}$;
calibration coefficient $\lambda$.

\State Initialize the category-conditional Gaussian parameters
$\{\pi_{m,c}^{0},
\boldsymbol{\mu}_{m,c}^{0},
\boldsymbol{\Sigma}_{m,c}^{0}\}$
from the pretrained classifier according to
Eqs.~\eqref{eq:init_mu_cov} and~\eqref{eq:init_pi},
for $m\in\{m_1,m_2,F\}$ and $c=1,\ldots,C$.

\State Initialize
$N_{m,c}^{0}\leftarrow 0$,
$\boldsymbol{S}_{m,c}^{0}\leftarrow\boldsymbol{0}$, and
$\boldsymbol{Q}_{m,c}^{0}\leftarrow\boldsymbol{0}$
for $m\in\{m_1,m_2,F\}$ and $c=1,\ldots,C$.

\State Initialize
$\boldsymbol{Q}_{Fm,c}^{0}\leftarrow\boldsymbol{0}$
and
$\boldsymbol{\Sigma}_{Fm,c}^{0}\leftarrow\boldsymbol{I}$
for $m\in\{m_1,m_2\}$ and $c=1,\ldots,C$.

\State Collect all initialized sufficient statistics and Gaussian parameters
into the online state $\mathcal{S}^{0}$.

\For{$t=1$ to $T$}
    \State Partition $\mathcal{D}^{t}$ according to
    Eq.~\eqref{eq:batch-partition}, and extract all available
    unimodal/multimodal representations and source logits $\boldsymbol s_i^t$.

    \State Compute the soft responsibilities from $\boldsymbol{s}_{i}^{t}$,
    and update the online state
    $\mathcal{S}^{t-1}\rightarrow\mathcal{S}^{t}$
    using the incremental sufficient statistics and EMA updates.

    \State Construct the Gaussian calibration logits
    $\boldsymbol r_i^t$ using
    Eq.\eqref{eq:gaussian-calibration-logits}.

    \State Compute the Gaussian prediction alignment loss
    $\mathcal{L}_g^t$ using Eq.~\eqref{eq:calibration-loss}.

    \State For complete samples, detect the unreliable modality by
    symmetric KL divergence and compute the one-sided rectification loss
    $\mathcal{L}_c^t$ using Eq.~\eqref{eq:onesided_infonce_instance};
    set $\mathcal{L}_c^t=0$ if
    $\mathcal{D}_{\mathrm{full}}^t=\varnothing$.

    \State Update the adaptable model parameters:
    \[
    \theta^{t}
    \gets
    \theta^{t-1}
    -\eta\nabla_{\theta}
    \left(w_\text{g}\mathcal{L}_\text{g}^t+w_\text{c}\mathcal{L}_\text{c}^t\right).
    \]

    \State Obtain the calibrated logits using
    Eq.~\eqref{eq:calibrated-logits}, and predict
    \[
    \widehat y_i^t
    =\arg\max_c\widetilde{s}_{i,c}^t.
    \]

    \State \textbf{Output}
    $(\widehat{\boldsymbol y}^{t},\theta^{t},\mathcal{S}^{t})$;
    pass $(\theta^{t},\mathcal{S}^{t})$ to batch $t+1$.
\EndFor

\State \Return.
\end{algorithmic}
\end{algorithm*}

\subsection{Gaussian-Guided Prediction Calibration}

Based on the online category-conditional Gaussian statistics, we construct Gaussian calibration logits to capture the evolving category structure of the target stream. 
For each incoming sample, we construct the Gaussian calibration logits $\boldsymbol{r}_{i}^{t}$ based on the availability of its multimodal representations, distinguishing between modality-complete and modality-incomplete samples.

For a complete sample $i\in\mathcal{D}_{\mathrm{full}}^{t}$, the fusion representation $\boldsymbol{z}_{F,i}^{t}$ is directly observable. 
Since the fusion representation integrates information from all available modalities, its category-conditional Gaussian model provides a more comprehensive description of the sample's category-specific distribution.
We therefore obtain its Gaussian calibration logits from the category-conditional Gaussian model associated with the fusion perspective as
\begin{equation}
\boldsymbol{r}_{i}^{t}
=
\boldsymbol{g}_{F}
\left(
\boldsymbol{z}_{F,i}^{t}
\right).
\end{equation}
These logits characterize the category preference induced by the updated Gaussian distribution and are directly used as calibration signals for complete samples.

For an incomplete sample $i\in\mathcal{D}_{m}^{t}$, where $m\in\{m_1,m_2\}$ denotes the observed modality, only the modality-specific representation $\boldsymbol{z}_{m,i}^{t}$ is available, while the fusion representation cannot be directly obtained.
Therefore, based on the category-conditional Gaussian modeling of the observed modality, we can only obtain the log-posterior induced by the available representation, given by
$
\boldsymbol{g}_{m}
\left(
\boldsymbol{z}_{m,i}^{t}
\right),
$
while the prediction induced by the unavailable fusion representation, which exploits the complete multimodal representation, cannot be directly obtained.

To incorporate the unavailable fusion representation into prediction, we estimate its category-conditioned value for each candidate category \(k\) using the learned cross-perspective statistics.
Since the observed modality representation and the fusion representation are jointly modeled by a category-conditional Gaussian distribution, the conditional distribution of the fusion representation given \(\boldsymbol{z}_{m,i}^{t}\) also follows  Gaussian distribution. We therefore use its conditional mean as a point estimate:
\begin{equation}
\hat{\boldsymbol{z}}_{F\mid m,k,i}^{t}
=
\boldsymbol{\mu}_{F,k}^{t}
+
\boldsymbol{\Sigma}_{Fm,k}^{t}
\left(
\boldsymbol{\Sigma}_{m,k}^{t}
\right)^{-1}
\left(
\boldsymbol{z}_{m,i}^{t}
-
\boldsymbol{\mu}_{m,k}^{t}
\right).
\end{equation}
This formulation expresses the latent fusion representation through a collection of category-conditioned estimates $\hat{\boldsymbol{z}}_{F\mid m,k,i}^{t}$, with each estimate corresponding to the conditional expectation associated with category $k$.

Since the category condition is unknown, we consider all candidate categories when estimating the fusion representation. 
The posterior prediction from the observed modality is then used to weight the fusion prediction obtained under each category condition.
Specifically, the weight for candidate category $k$ is given by
$
\operatorname{Softmax}
\left(
\boldsymbol{g}_{m}
\left(
\boldsymbol{z}_{m,i}^{t}
\right)
\right)_{k}.
$ 
For each candidate category \(k\), the corresponding estimate \(\hat{\boldsymbol{z}}_{F\mid m,k,i}^{t}\) is evaluated by the Gaussian model of the fusion perspective, yielding the prediction score for each output category. We aggregate these prediction scores across all candidate category conditions using the posterior weights from the observed modality. Accordingly, the recovered fusion logit for category \(c\) is defined as
\begin{equation}
\label{eq:recovery-logit}
g_{m\rightarrow F,c}
\left(
\boldsymbol{z}_{m,i}^{t}
\right)
=
\sum_{k=1}^{C}
\operatorname{Softmax}
\left(
\boldsymbol{g}_{m}
\left(
\boldsymbol{z}_{m,i}^{t}
\right)
\right)_{k}
\cdot
g_{F,c}
\left(
\hat{\boldsymbol{z}}_{F\mid m,k,i}^{t}
\right).
\end{equation}
Here, $k$ specifies the category condition used to estimate the latent fusion representation, whereas $c$ denotes the output category of the resulting recovery logits. The corresponding vector of recovery logits is denoted by
$
\boldsymbol{g}_{m\rightarrow F}
\left(
\boldsymbol{z}_{m,i}^{t}
\right)
=
\left[
g_{m\rightarrow F,1}
\left(
\boldsymbol{z}_{m,i}^{t}
\right),
\ldots,
g_{m\rightarrow F,C}
\left(
\boldsymbol{z}_{m,i}^{t}
\right)
\right]^{\top}.
$
The probabilistic formulation and the rationale for the aggregation rule are provided in Appendix ~\ref{appendix:recovery-derivation}.

The modality-specific log-posterior and the recovery logits provide complementary contributions to the Gaussian calibration of an incomplete sample.
The former is evaluated directly from the observed modality, whereas the latter is derived from the estimated fusion representation.
We therefore combine the two terms with equal weighting to obtain the Gaussian calibration logits for incomplete samples as
\begin{equation}
\boldsymbol{r}_{i}^{t}
=
\frac{1}{2}
\left[
\boldsymbol{g}_{m}
\left(
\boldsymbol{z}_{m,i}^{t}
\right)
+
\boldsymbol{g}_{m\rightarrow F}
\left(
\boldsymbol{z}_{m,i}^{t}
\right)
\right].
\label{eq:equal-weighting}
\end{equation}
The effect of the relative weighting between the two terms is examined in Section \ref{sec:exp-further}.

Combining the complete and incomplete cases, the Gaussian calibration logits for each incoming sample are defined as
\begin{equation}
\label{eq:gaussian-calibration-logits}
\boldsymbol{r}_{i}^{t}
=
\begin{cases}
\boldsymbol{g}_{F}
\left(
\boldsymbol{z}_{F,i}^{t}
\right),
&
i\in\mathcal{D}_{\mathrm{full}}^{t},
\\[6pt]
\dfrac{1}{2}
\left[
\boldsymbol{g}_{m}
\left(
\boldsymbol{z}_{m,i}^{t}
\right)
+
\boldsymbol{g}_{m\rightarrow F}
\left(
\boldsymbol{z}_{m,i}^{t}
\right)
\right],
&
i\in\mathcal{D}_{m}^{t}.
\end{cases}
\end{equation}
Accordingly, complete samples are calibrated directly from the Gaussian evidence of their fusion representations, while incomplete samples combine the evidence available from the observed modality with that inferred for the latent fusion representation.

The Gaussian calibration logits $\boldsymbol{r}_{i}^{t}$ provide distributional evidence derived from the online category-conditional statistics, whereas the source logits $\boldsymbol{s}_{i}^{t}$ retain the discriminative knowledge learned from the source data. However, the two predictions may exhibit distributional discrepancy, which can weaken their complementarity when directly combined. We therefore align the source model prediction with the Gaussian calibration prediction before performing logit-level fusion.

Specifically, we convert the Gaussian calibration logits into a soft prediction and use it as a fixed reference to supervise the source prediction. The resulting soft-prediction alignment loss is defined as
\begin{equation}
\label{eq:calibration-loss}
\mathcal{L}_\text{g}
=
-
\frac{1}{\lvert\mathcal{B}\rvert}
\sum_{i\in\mathcal{B}}
\sum_{c=1}^{C}
\operatorname{Softmax}
\left(
\boldsymbol{r}_{i}^{t}
\right)
_{c}
\log
\big (
\operatorname{Softmax}
\left(
\boldsymbol{s}_{i}^{t}
\right)
_{c}
\big ).
\end{equation}

After alignment, we combine the source logits and the Gaussian calibration logits as
\begin{equation}
\label{eq:calibrated-logits}
\tilde{\boldsymbol{s}}_{i}^{t}
=
\boldsymbol{s}_{i}^{t}
+
\lambda\boldsymbol{r}_{i}^{t},
\end{equation}
where $\lambda$ controls the contribution of the Gaussian calibration logits to the final prediction.

In this way, the soft-prediction alignment progressively adapts the source model toward the category structure captured from the test stream, while the logit-level fusion directly integrates the discriminative source prediction with the online Gaussian evidence.

\subsection{Reliability-Aware Representation Rectification}

To address single-modality reliability shifts, we selectively rectify the representation of the degraded modality using information from the relatively reliable modality. 
Since both unimodal and multimodal representations are required for reliability assessment and cross-modal alignment, this rectification is performed only on complete-modality test instances in $\mathcal{D}_{\mathrm{full}}^t$.
Specifically, we first identify the degraded modality for each test sample according to the discrepancy between unimodal and multimodal predictions, and then perform one-sided contrastive alignment to rectify its representation while preserving the reliable modality.

From the log-posterior, the probabilistic prediction for each representation $m\in\{m_1,m_2,F\}$ is given by
$p(c\mid\boldsymbol{z}_{m,i}^t)=\text{Softmax}(\boldsymbol{g}_{m}(\boldsymbol{z}_{m,i}^t))_c$.
As demonstrated in \cite{guo2025smoothing}, the multimodal prediction $p(c\mid \boldsymbol{z}_{F,i}^t)$ is generally more robust and stable than the unimodal predictions $p(c\mid \boldsymbol{z}_{m_1,i}^t)$ and $p(c\mid \boldsymbol{z}_{m_2,i}^t)$.
We therefore use the multimodal prediction as a reference, such that a modality whose prediction exhibits a larger distribution discrepancy from it is considered more likely to have undergone distribution shift.

Therefore, we detect modalities exhibiting distribution shift by comparing the distribution discrepancies between different unimodal predictions and multimodal predictions. Specifically, we compute the distribution discrepancy by
\begin{equation}
\begin{aligned}
& D_{m_1,i}^t=D_\text{SKL}\big(p(c\mid\boldsymbol{z}_{m_1,i}^t),p(c\mid \boldsymbol{z}_{F,i}^t)\big), \\
& D_{m_2,i}^t=D_\text{SKL}\big(p(c\mid\boldsymbol{z}_{m_2,i}^t),p(c\mid \boldsymbol{z}_{F,i}^t)\big),
\end{aligned}
\end{equation}
where $D_\text{SKL}(\cdot,\cdot)$ is the symmetric KL divergence term calculated by
$
D_\text{SKL}(p,q)=\frac{1}{2} \Big( D_\text{KL}(p\Vert q)+D_\text{KL}(q\Vert p)\Big). 
$
Based on these discrepancies, the complete-modality test instances are partitioned according to which modality is less reliable:
\begin{equation}
\begin{aligned}
\mathcal{R}_{m_1}^t
&=
\left\{
i \;\middle|\;
i \in \mathcal{D}_{\mathrm{full}}^t,\;
D_{m_1,i}^t > D_{m_2,i}^t
\right\}, \\
\mathcal{R}_{m_2}^t
&=
\left\{
i \;\middle|\;
i \in \mathcal{D}_{\mathrm{full}}^t,\;
D_{m_2,i}^t \ge D_{m_1,i}^t
\right\}.
\end{aligned}
\end{equation}
Here, $\mathcal{R}_{m_1}^t$ and $\mathcal{R}_{m_2}^t$ denote the index sets of complete-modality test instances for which modalities $m_1$ and $m_2$ are identified as the less reliable modality, respectively.

Therefore, we align the representation of the less reliable modality with that of the reliable modality using a one-sided InfoNCE loss. Gradients propagate only through the less reliable modality representation, while gradients through the reliable modality representation are stopped to prevent the unreliable modality from altering it.
Let $\bar{\boldsymbol{z}}_{m_1,i}^t=\boldsymbol{z}_{m_1,i}^t/\|\boldsymbol{z}_{m_1,i}^t\|_2$  and $\bar{\boldsymbol{z}}_{m_2,i}^t=\boldsymbol{z}_{m_2,i}^t/\|\boldsymbol{z}_{m_2,i}^t\|_2$ 
be $\ell_2$-normalized representations and $\tau=0.05$ be the temperature.
We denote stop-gradient by $\operatorname{sg}(\cdot)$ and inner product similarity by $\langle \cdot,\cdot\rangle$. For each sample $i$ in batch $t$, the loss is defined as
\begin{equation}
\ell_i =
\begin{cases}
-\log
\dfrac{
\exp\!\left(
\left\langle
\bar{z}_{m_1,i}^{t},
\operatorname{sg}(\bar{z}_{m_2,i}^{t})
\right\rangle/\tau
\right)
}{
\displaystyle
\sum_{\mathclap{j\in\mathcal{D}_{\mathrm{full}}^t}}
\exp\!\left(
\left\langle
\bar{z}_{m_1,i}^{t},
\operatorname{sg}(\bar{z}_{m_2,j}^{t})
\right\rangle/\tau
\right)
},
& i\in\mathcal{R}_{m_1}^{t},
\\[1.5em]
-\log
\dfrac{
\exp\!\left(
\left\langle
\bar{z}_{m_2,i}^{t},
\operatorname{sg}(\bar{z}_{m_1,i}^{t})
\right\rangle/\tau
\right)
}{
\displaystyle
\sum_{\mathclap{j\in\mathcal{D}_{\mathrm{full}}^t}}
\exp\!\left(
\left\langle
\bar{z}_{m_2,i}^{t},
\operatorname{sg}(\bar{z}_{m_1,j}^{t})
\right\rangle/\tau
\right)
},
& i\in\mathcal{R}_{m_2}^{t}.
\end{cases}
\label{eq:onesided_infonce_instance}
\end{equation}

The overall alignment loss at batch $t$ is defined as the average of the sample-wise losses over all complete-modality test instances:
$
\mathcal{L}_{c} = \frac{1}{|\mathcal{D}^{t}_{\mathrm{full}}|}\sum_{i\in \mathcal{D}^{t}_{\mathrm{full}}} \ell_i.
$

The overall loss is
\begin{equation}
\mathcal{L}
=
w_{\text{g}}\,\mathcal{L}_{\text{g}}
+
w_{\text{c}}\,\mathcal{L}_{\text{c}},
\end{equation}
where only $\mathcal{L}_{\text{g}}$ and $\mathcal{L}_{\text{c}}$ are weighted by hyperparameters $(w_{\text{g}},w_{\text{c}})$.

\begin{table*}[!t]
\caption{
Prediction accuracies (in \%) on the Kinetics50-C benchmark with corrupted video inputs.
}
\label{tab:ks50_video_c}
\centering
% \scriptsize
\begin{adjustbox}{width=\textwidth}
\begin{tabular}{l*{16}{c}}
\toprule
\multirow{2}{*}{Models}
& \multicolumn{3}{c}{Noise}
& \multicolumn{4}{c}{Blur}
& \multicolumn{4}{c}{Weather}
& \multicolumn{4}{c}{Digital}
& \multirow{2}{*}{Avg.} \\
\cmidrule(lr){2-4}
\cmidrule(lr){5-8}
\cmidrule(lr){9-12}
\cmidrule(lr){13-16}
& Gauss. & Shot & Impul.
& Defoc. & Glass & Mot. & Zoom
& Snow & Frost & Fog & Brit.
& Contr. & Elas. & Pix. & JPEG & \\
\midrule

Source
& 46.8 & 48.0 & 46.9
& 67.5 & 62.2 & 70.8 & 66.7
& 61.6 & 60.3 & 46.7 & 75.2
& 52.1 & 65.7 & 66.5 & 61.9 & \cellcolor{yellow!40}59.9 \\

Tent\venue{ICLR 2021}
& 46.3 & 47.0 & 46.3
& 67.2 & 62.5 & 71.0 & 67.6
& 63.1 & 61.1 & 34.9 & 75.4
& 51.6 & 66.8 & 67.2 & 62.7 & \cellcolor{yellow!40}59.4 \\

EATA\venue{ICML 2022}
& 46.8 & 47.6 & 47.1
& 67.2 & 62.7 & 70.6 & 67.2
& 62.3 & 60.9 & 46.7 & 75.2
& 52.4 & 65.9 & 66.8 & 62.5 & \cellcolor{yellow!40}60.1 \\

SAR\venue{ICLR 2023}
& 46.7 & 47.4 & 46.8
& 67.0 & 61.9 & 70.4 & 66.4
& 61.8 & 60.6 & 46.0 & 75.2
& 52.1 & 65.7 & 66.4 & 62.0 & \cellcolor{yellow!40}59.8 \\

MMT\venue{CVPR 2022}
& 46.2 & 46.6 & 46.1
& 58.8 & 55.7 & 62.4 & 61.7
& 52.6 & 54.4 & 48.5 & 69.3
& 49.3 & 57.6 & 56.4 & 54.5 & \cellcolor{yellow!40}54.5 \\

READ\venue{ICLR 2024}
& 49.4 & 49.7 & 49.0
& 68.0 & 65.1 & 71.2 & 69.0
& 64.5 & 64.4 & 57.4 & 75.5
& 53.6 & 68.3 & 68 & 65.1 & \cellcolor{yellow!40}62.5 \\

ABPEM\venue{AAAI 2025}
& 50.3 & 51.1 & 50.4
& 70.0 & 69.6 & \textbf{72.5} & 71.2
& 65.2 & 66.2 & 65.6 & 75.7
& 56.6 & 71.9 & 70.5 & 67.8 & \cellcolor{yellow!40}65.0 \\

SuMi\venue{ICLR 2025}
& 50.1 & 50.7 & 50.4
& 68.2 & 65.6 & 72.2 & 69.7
& 65.7 & 67.0 & 56.5 & \textbf{77.1}
& 55.2 & 69.3 & 71.2 & 68.9 & \cellcolor{yellow!40}63.9 \\

TSA\venue{ICML 2025}
& 52.6 & 52.3 & 52.0
& 68.7 & 68.0 & 70.7 & 68.8
& 65.2 & 66.6 & 64.3 & 74.6
& 57.4 & 70.5 & 69.0 & 66.2 & \cellcolor{yellow!40}64.5 \\

BriMPR\venue{AAAI 2026}
&\textbf{55.3}	&\textbf{56.1}	&\textbf{56.7}	
&67.8	&67.9	&70.6	&68.8	
&65.9	&66.2	&64.1	&76.2	
&56.3	&72.0	&\textbf{73.7}	&70.5	&\cellcolor{yellow!40}65.9 \\

\rowcolor{blue!20}
\textbf{AdaPGC}\venue{This Work}
& 52.2 & 54.4 & 53.9
& \textbf{70.7} & \textbf{70.9} & 72.4 & \textbf{72.2}
& \textbf{67.1} & \textbf{68.8} & \textbf{67.4} & 76.4
& \textbf{58.0} & \textbf{72.9} & 73.1 & \textbf{70.7} &  \textbf{66.7} \\

\bottomrule
\end{tabular}
\end{adjustbox}
\end{table*}

\begin{table*}[t]
\caption{
Prediction accuracies (in \%) on the VGGSound-C benchmark with corrupted video inputs.
}
\label{tab:vgg_video_c}
\centering
% \scriptsize
\begin{adjustbox}{width=\textwidth}
\begin{tabular}{lccccccccccccccccc}
\toprule
\multirow{2}{*}{Models}
& \multicolumn{3}{c}{Noise}
& \multicolumn{4}{c}{Blur}
& \multicolumn{4}{c}{Weather}
& \multicolumn{4}{c}{Digital}
& \multirow{2}{*}{Avg.} \\
\cmidrule(lr){2-4}
\cmidrule(lr){5-8}
\cmidrule(lr){9-12}
\cmidrule(lr){13-16}
& Gauss. & Shot & Impul.
& Defoc. & Glass & Mot. & Zoom
& Snow & Frost & Fog & Brit.
& Contr. & Elas. & Pix. & JPEG \\
\midrule

Source
& 52.8 & 52.7 & 52.7
& 57.2 & 57.2 & 58.7 & 57.6
& 56.4 & 56.6 & 55.6 & 58.9
& 53.7 & 56.9 & 55.8 & 56.9 & \cellcolor{yellow!40}56.0 \\

Tent\venue{ICLR 2021}
& 52.7 & 52.7 & 52.7
& 56.7 & 56.5 & 57.9 & 57.2
& 55.9 & 56.3 & 56.3 & 58.4
& 54.0 & 57.4 & 56.2 & 56.7 & \cellcolor{yellow!40}55.8 \\

EATA\venue{ICML 2022}
& 53.0 & 52.8 & 53.0
& 57.2 & 57.1 & 58.6 & 57.8
& 56.3 & 56.8 & 56.4 & 59.0
& 54.1 & 57.4 & 56.1 & 57.0 & \cellcolor{yellow!40}56.2 \\

SAR\venue{ICLR 2023}
& 52.9 & 52.8 & 52.9
& 57.2 & 57.1 & 58.6 & 57.6
& 56.3 & 56.7 & 55.9 & 58.9
& 54.0 & 57.0 & 56.0 & 57.0 & \cellcolor{yellow!40}56.1 \\

MMT\venue{CVPR 2022}
& 7.1 & 7.3 & 7.3
& 44.8 & 41.5 & 48.0 & 45.5
& 27.4 & 23.5 & 30.5 & 46.3
& 24.0 & 43.0 & 40.7 & 45.7 & \cellcolor{yellow!40}32.0 \\

READ\venue{ICLR 2024}
& 53.6 & 53.6 & 53.5
& 57.9 & 57.7 & 59.4 & 58.8
& 57.2 & 57.8 & 55.0 & 59.9
& 55.2 & 58.6 & 57.1 & 57.9 & \cellcolor{yellow!40}56.9 \\

ABPEM\venue{AAAI 2025}
& 54.0 & 53.9 & 54.0
& \textbf{58.2} & 58.1 & \textbf{59.6} & \textbf{59.3}
& 57.5 & \textbf{58.2} & 58.2 & 60.2
& \textbf{56.2} & 59.1 & 57.5 & 58.3
& \cellcolor{yellow!40}57.5 \\

SuMi\venue{ICLR 2025}
& 54.0 & 54.3 & 53.8
& \textbf{58.2} & 58.4 & 59.4 & 58.7
& 57.5 & \textbf{58.2} & 57.6 & 59.4
& 54.8 & 59.0 & 57.5 & 58.2
& \cellcolor{yellow!40}57.3 \\

TSA\venue{ICML 2025}
& 53.9 & 53.9 & 54.0
& 57.6 & 58.0 & 59.0 & 58.6
& 56.9 & 57.0 & 56.6 & 59.8
& 54.7 & 58.6 & 56.7 & 57.9 & \cellcolor{yellow!40}56.9 \\

BriMPR\venue{AAAI 2026}
& \textbf{54.9} & \textbf{55.0} & 55.0
& 57.9 & \textbf{58.5} & 58.9 & 58.7
& 57.5 & 58.0 & \textbf{58.5} & \textbf{60.3}
& 54.5 & \textbf{59.7} & \textbf{59.3} & \textbf{59.0}
& \cellcolor{yellow!40}\textbf{57.7} \\

\rowcolor{blue!20}
\textbf{AdaPGC}\venue{This Work}
&\textbf{54.9}	&54.9	&\textbf{55.1}	&\textbf{58.2}	&58.2	&59.5	&58.9	&\textbf{57.6}	&57.8	&\textbf{58.5}	&59.8	&55.2	&59.2	&58.4	&58.6	&\textbf{57.7}\\

\bottomrule
\end{tabular}
\end{adjustbox}
\end{table*}

\begin{table*}[t]
\caption{
Prediction accuracies (in \%) on the Kinetics50-C (left) and VGGSound-C (right) benchmarks with corrupted audio inputs.
}
\label{tab:audio_c}
\centering
\scriptsize
\begin{adjustbox}{width=\textwidth}
\begin{tabular}{l*{14}{c}}
\toprule
\multirow{2}{*}{Models}
& \multicolumn{3}{c}{Noise}
& \multicolumn{3}{c}{Weather}
& \multirow{2}{*}{Avg.}
& \multicolumn{3}{c}{Noise}
& \multicolumn{3}{c}{Weather}
& \multirow{2}{*}{Avg.} \\
\cmidrule(lr){2-4}
\cmidrule(lr){5-7}
\cmidrule(lr){9-11}
\cmidrule(lr){12-14}
& Gauss. & Traff. & Crowd.
& Rain & Thund. & Wind &
& Gauss. & Traff. & Crowd.
& Rain & Thund. & Wind & \\
\midrule

Source
& 73.7 & 65.5 & 67.9 & 70.3 & 67.9 & 70.3 & \cellcolor{yellow!40}69.3
& 37.0 & 25.5 & 16.8 & 21.6 & 27.3 & 25.5 & \cellcolor{yellow!40}25.6 \\

Tent\venue{ICLR 2021}
& 73.9 & 67.4 & 69.2 & 70.4 & 66.5 & 70.5 & \cellcolor{yellow!40}69.6
& 10.6 & 2.6 & 1.8 & 2.8 & 5.3 & 4.1 & \cellcolor{yellow!40}4.5 \\

EATA\venue{ICML 2022}
& 73.7 & 66.1 & 68.5 & 70.3 & 67.9 & 70.1 & \cellcolor{yellow!40}69.4
& 39.2 & 26.1 & 22.9 & 26.0 & 31.7 & 30.4 & \cellcolor{yellow!40}29.4 \\

SAR\venue{ICLR 2023}
& 73.7 & 65.4 & 68.2 & 69.9 & 67.2 & 70.2 & \cellcolor{yellow!40}69.1
& 37.4 & 9.5 & 11.0 & 12.1 & 26.8 & 23.7 & \cellcolor{yellow!40}20.1 \\

MMT\venue{CVPR 2022}
& 70.8 & 69.2 & 68.5 & 69.0 & 69.8 & 68.5 & \cellcolor{yellow!40}69.4
& 14.1 & 5.2 & 6.4 & 9.8 & 8.6 & 4.5 & \cellcolor{yellow!40}7.6 \\

READ\venue{ICLR 2024}
& 74.1 & 69.0 & 69.7 & 71.1 & 71.8 & 70.7 & \cellcolor{yellow!40}71.1
& 40.4 & 28.9 & 26.6 & 30.9 & 36.7 & 30.6 & \cellcolor{yellow!40}32.4 \\

ABPEM\venue{AAAI 2025}
& 74.8 & 71.3 & 71.5
& 71.9 & 73.8 & 71.6
& \cellcolor{yellow!40}72.5
& 40.6 & 33.7 & 34.8
& 32.2 & 41.1 & 34.4
& \cellcolor{yellow!40}36.1 \\

SuMi\venue{ICLR 2025}
& 75.1 & 68.9 & 70.6
& 71.6 & 72.8 & 72.1
& \cellcolor{yellow!40}71.9
& \textbf{41.9} & 26.3 & 27.9
& 31.6 & 37.1 & 34.1
& \cellcolor{yellow!40}33.2 \\

TSA\venue{ICML 2025}
& 74.5 & 69.6 & 70.5
& 71.4 & 72.0 & 71.0
& \cellcolor{yellow!40}71.5
& 41.5 & 31.8 & 30.9
& 32.6 & 38.9 & 32.6
& \cellcolor{yellow!40}34.7 \\

BriMPR\venue{AAAI 2026}
& 74.8 & 69.6 & 71.7
& 71.5 & 72.4 & 72.0
& \cellcolor{yellow!40}72.0
& 39.3 & \textbf{35.0} & 36.7
& 32.5 & 41.0 & \textbf{34.6}
& \cellcolor{yellow!40}36.5 \\

\rowcolor{blue!20}
\textbf{AdaPGC}\venue{This Work}
& \textbf{75.3} & \textbf{72.1} & \textbf{72.7}
& \textbf{72.2} & \textbf{75.4} & \textbf{72.4}
& \textbf{73.3}
& 41.1 & 34.3 & \textbf{37.2}
& \textbf{34.1} & \textbf{43.7} & 34.4
& \textbf{37.5} \\

\bottomrule
\end{tabular}
\end{adjustbox}
\end{table*}

\begin{table*}[t]
\centering
\scriptsize
\caption{
Prediction accuracies (in \%) on the Kinetics50 benchmark under modality-missing scenarios.
}
\label{tab:missing_ks50}
\begin{adjustbox}{width=\textwidth}
\begin{tabular}{l*{16}{c}}
\toprule
\multirow{2}{*}{Method}
& \multicolumn{5}{c}{Missing Audio Ratio}
& \multicolumn{5}{c}{Missing Video Ratio}
& \multicolumn{5}{c}{Missing Both Ratio}
& \multirow{2}{*}{\makecell{Avg.}} \\

\cmidrule(lr){2-6}
\cmidrule(lr){7-11}
\cmidrule(lr){12-16}

& 50\% & 60\% & 70\% & 80\% & 90\%
& 50\% & 60\% & 70\% & 80\% & 90\%
& 50\% & 60\% & 70\% & 80\% & 90\%
& \\
\midrule

Source
& 74.9 & 72.6 & 72.3 & 70.0 & 68.8
& 64.9 & 61.0 & 58.3 & 53.8 & 50.5
& 70.0 & 68.3 & 65.5 & 62.8 & \underline{59.5}
& \cellcolor{yellow!40}64.9 \\

Tent\venue{ICLR 2021}
&75.0	&72.8	&72.7	&70.1	&69.0	
&65.2	&61.6	&58.2	&53.7	&50.4	
&70.4	&68.5	&66.0	&62.9	&59.4	
&\cellcolor{yellow!40}65.1 \\

EATA\venue{ICML 2022}
&75.5	&72.7	&72.5	&70.5	&69.2	
&65.3	&61.3	&58.4	&53.8	&50.9	
&70.2	&68.7	&65.8	&63.1	&59.2	
&\cellcolor{yellow!40}65.1 \\

SAR\venue{ICLR 2023}
&75.1	&73.0	&72.5	&70.3	&68.9	
&64.9	&61.1	&58.2	&53.6	&50.4	
&70.2	&68.5	&65.4	&62.9	&59.5	
&\cellcolor{yellow!40}65.0 \\

MMT\venue{CVPR 2022}
&71.5	&71.0	&70.4	&69.3	&68.2
&62.0	&58.7	&56.2	&52.8	&49.9	
&67.0	&66.3	&63.6	&61.6	&59.1	
&\cellcolor{yellow!40}63.2 \\

READ\venue{ICLR 2024}
& 74.7 & 73.3 & 72.0 & 70.1 & 69.2
& 64.7 & 61.1 & 56.8 & 53.0 & 49.3
& 70.5 & 68.0 & 64.5 & 61.4 & 57.4
& \cellcolor{yellow!40}64.4 \\

ABPEM\venue{AAAI 2025}
& 75.2 & 73.0 & 72.9 & 71.0 & 69.3
& 65.6 & 61.9 & 58.3 & 54.1 & 50.6
& 70.4 & 68.7 & 65.6 & 62.9 & 59.4
& \cellcolor{yellow!40}65.3 \\

SuMi\venue{ICLR 2025}
& 75.2 & 72.6 & 72.5 & 70.2 & 69.0
& 65.1 & 61.3 & 58.2 & 53.7 & 50.8
& 70.1 & 68.6 & 65.7 & 62.9 & 59.4
& \cellcolor{yellow!40}65.0 \\

TSA\venue{ICML 2025}
& 73.9 & 72.5 & 72.3 & 69.8 & 69.4
& 64.4 & 60.7 & 58.7 & 54.5 & 50.6
& 68.8 & 66.1 & 65.4 & 62.0 & 59.0
& \cellcolor{yellow!40}64.5 \\

BriMPR\venue{AAAI 2026}
&75.0	&73.7	&72.0	&70.9	&69.8	
&64.3	&60.0	&57.3	&53.2	&50.0
&70.3	&67.2	&65.0	&61.8	&57.8	
&\cellcolor{yellow!40}64.5
 \\

\rowcolor{blue!20}\textbf{AdaPGC}\venue{This Work}
& \textbf{75.5} & \textbf{73.9} & \textbf{73.4} & \textbf{71.8} & \textbf{70.7}
& \textbf{65.8} & \textbf{62.5} & \textbf{59.9} & \textbf{55.5} & \textbf{52.2}
& \textbf{70.7} & \textbf{69.4} & \textbf{66.3} & \textbf{63.6} & \textbf{61.1}
& \textbf{66.2}

\\
    \bottomrule
\end{tabular}
\end{adjustbox}
\end{table*}

\begin{table*}[!t]
\centering
\scriptsize
\caption{
Prediction accuracies (in \%) on the VGGSound benchmark under modality-missing scenarios.
}
\label{tab:missing_vgg}
\begin{adjustbox}{width=\textwidth}
\begin{tabular}{l*{16}{c}}
\toprule
\multirow{2}{*}{Method}
& \multicolumn{5}{c}{Missing Audio Ratio}
& \multicolumn{5}{c}{Missing Video Ratio}
& \multicolumn{5}{c}{Missing Both Ratio}
& \multirow{2}{*}{\makecell{Avg.}} \\

\cmidrule(lr){2-6}
\cmidrule(lr){7-11}
\cmidrule(lr){12-16}

& 50\% & 60\% & 70\% & 80\% & 90\%
& 50\% & 60\% & 70\% & 80\% & 90\%
& 50\% & 60\% & 70\% & 80\% & 90\%
& \\
\midrule
        
Source
& \textbf{46.7} & \textbf{43.2} & 39.4 & 36.6 & 32.8
& 58.0 & 56.5 & 55.5 & 54.2 & 53.3
& \textbf{52.1} & 50.0 & 47.3 & 45.1 & 43.2
& \cellcolor{yellow!40}47.6 \\

Tent\venue{ICLR 2021}
&42.9	&38.1	&32.2	&27.1	&22.1	
&57.0	&55.7	&54.9	&53.6	&52.7	
&50.9	&48.2	&45.6	&43.1	&41.0	
&\cellcolor{yellow!40}44.3 \\

EATA\venue{ICML 2022}
&45.3	&42.1	&38.0	&35.3	&31.4	
&57.9	&56.5	&55.5	&54.3	&53.5	
&51.7	&49.3	&46.8	&44.6	&42.8	
&\cellcolor{yellow!40}47.0 \\
SAR	\venue{ICLR 2023}
&44.5	&40.7	&36.8	&33.7	&29.3	
&55.3	&53.9	&52.6	&50.9	&49.8	
&50.0	&47.6	&44.8	&42.8	&40.7	
&\cellcolor{yellow!40}44.9 \\

MMT\venue{CVPR 2022}
&43.4	&40.5	&37.5	&34.8	&31.8	
&55.1	&54.5	&54.0	&53.5	&53.1	
&49.1	&47.5	&45.7	&43.7	&42.6	
&\cellcolor{yellow!40}45.8 \\

READ\venue{ICLR 2024}
& 43.9 & 40.2 & 35.1 & 30.5 & 18.0
& 56.9 & 55.4 & 54.3 & 53.2 & 51.7
& 50.4 & 47.9 & 45.2 & 42.7 & 40.5
& \cellcolor{yellow!40}44.4 \\

ABPEM\venue{AAAI 2025}
& 42.2 & 39.5 & 36.1 & 33.9 & 31.1
& 52.4 & 51.5 & 51.0 & 50.4 & 50.5
& 46.9 & 45.2 & 43.3 & 41.6 & 41.0
& \cellcolor{yellow!40}43.8 \\

SuMi\venue{ICLR 2025}
& 46.0 & 42.5 & 38.8 & 36.1 & 32.3
& 58.0 & 56.4 & 55.4 & 54.4 & 53.4
& 51.8 & 49.7 & 47.0 & 44.8 & 43.0
& \cellcolor{yellow!40}47.3 \\

TSA\venue{ICML 2025}
& 42.7 & 40.1 & 35.3 & 33.4 & 29.4
& 56.1 & 54.4 & 53.5 & 52.3 & 50.1
& 48.6 & 46.8 & 44.5 & 42.7 & 39.2
& \cellcolor{yellow!40}44.6 \\

BriMPR\venue{AAAI 2026}
&44.0 &40.1 &36.1 &33.1  &29.0 &58.1 &56.8  &55.6 &54.4  &53.2 &50.3 &47.7  &44.6 &42.2 &40.0&\cellcolor{yellow!40}45.7\\

\rowcolor{blue!20}\textbf{AdaPGC}\venue{This Work}
& 46.3 & \textbf{43.2} & \textbf{39.6} & \textbf{36.8} & \textbf{33.6}
& \textbf{58.5} & \textbf{57.2} & \textbf{56.2} & \textbf{55.5} & \textbf{54.7}
& 52.0 & \textbf{50.2} & \textbf{47.7} & \textbf{46.1} & \textbf{44.3}
& \textbf{48.1} \\
\bottomrule
\end{tabular}
\end{adjustbox}
\end{table*}

\section{Experiments}

\subsection{Experimental Settings}

\paragraph{Datasets}
To evaluate robustness against different types of multimodal distribution shifts, we consider both modality corruption and modality missingness scenarios.
Specifically, we construct corrupted datasets, Kinetics50-C and VGGSound-C, derived from Kinetics50 \cite{kay2017kinetics} and VGGSound~\cite{chen2020vggsound}. 
Kinetics50 contains YouTube videos of human actions across 50 categories, while VGGSound includes 309 audio-visual categories, covering a wide range of real-world audio-visual events. 
Following previous work \cite{hendrycksbenchmarking,DBLP:conf/iclr/Yang0Z0024}, we introduce 15 visual distortions (e.g., Gaussian noise, motion blur, brightness variations) and six audio disturbances (e.g., traffic noise, rain noise) to simulate real-world distribution shifts. 
These distortions resulted in the corrupted Kinetics50-C and VGGSound-C datasets, which were then used for our evaluation, providing a challenging benchmark for multimodal test-time adaptation.
We further construct modality-missing test streams from the original Kinetics50 and VGGSound datasets to evaluate model robustness when one or more modalities become unavailable during inference.
Following previous work~\cite{lee2023multimodal,hu2024deep}, we consider three settings: 
Audio Missing, where $\eta\%$ of the samples contain only the visual modality; 
Visual Missing, where $\eta\%$ of the samples contain only the audio modality; and 
Both Missing, where $\frac{\eta}{2}\%$ of the samples are audio-only and another $\frac{\eta}{2}\%$ are visual-only. 
The remaining $(100-\eta)\%$ of the samples retain both modalities. We evaluate 5 missing rates, with $\eta \in \{50\%, 60\%, 70\%, 80\%, 90\%\}$.

\paragraph{Baselines}
We compare the proposed method AdaPGC with a comprehensive set of representative test-time adaptation approaches. 
Specifically, we include both unimodal and multimodal TTA methods to evaluate their effectiveness under challenging multimodal distribution shifts. 
For unimodal TTA, we consider entropy minimization-based methods, including TENT~\cite{wang2021tent}, EATA~\cite{pmlr-v162-niu22a}, and SAR~\cite{niu2023towards}. 
For multimodal TTA, we compare with MM-TTA~\cite{shin2022mm}, READ~\cite{DBLP:conf/iclr/Yang0Z0024}, ABPEM~\cite{zhao2025attention}, SuMi~\cite{guo2025smoothing}, BriMPR~\cite{li2026brimpr}, and TSA~\cite{chen2025testtime}, which are designed to exploit multimodal interactions and improve adaptation under modality variations. 
These baselines provide a comprehensive comparison with our probabilistic calibration framework.

\paragraph{Implementation Details}
For the source model, we follow READ \cite{DBLP:conf/iclr/Yang0Z0024} and adopt the pre-trained CAV-MAE \cite{gong2023contrastive}. 
All experiments use the Adam optimizer with a learning rate of $1\times10^{-4}$ and a batch size of 16 (Appendix~\ref{appendix:batchsize}). 
We configure the hyperparameters using loss weights $w_{\text{c}}=0.01$ and $w_{\text{g}}=1$. 
The fusion weight $\lambda$ is set to 1 for the modality corruption setting and 0.1 for the modality missingness setting. 
All experiments are conducted on a single NVIDIA A100 GPU.
During adaptation, we enforce a clear separation of update routes. 
The loss  $\mathcal{L}_{\text{g}}$  updates \emph{only} the fusion-layer attention parameters $W_{\Theta_h}$ and $B_{\Theta_h}$ for $h \in \{Q, K, V\}$, without modifying ny modality-specific encoders. In contrast, $\mathcal{L}_{\text{c}}$ updates \emph{only} the LayerNorm modules within the encoder of the modality flagged as corrupted by the reliability test.

\subsection{Main Results}

\begin{table*}[t]
\centering
\scriptsize
\caption{ 
Ablation study on the contributions of different components in AdaPGC on the Kinetics50 dataset. 
Three representative conditions are shown for each setting. 
Avg. is computed over all 15 Video-C corruptions, all 6 Audio-C corruptions, or all 15 missing-modality conditions.}
\label{tab:components_ablation}
\begin{adjustbox}{width=\textwidth}
\begin{tabular}{ccc|cccc|cccc|cccc}
\toprule
 & & & \multicolumn{4}{c|}{\textbf{Video-C}} & \multicolumn{4}{c|}{\textbf{Audio-C}} & \multicolumn{4}{c}{\textbf{Missing}} \\
 \textbf{FL} & \textbf{PA} & \textbf{AR} & Defoc. & Snow & Pix. & Avg. & Gauss. & Crowd & Wind & Avg. & A-70\% & V-70\% & Both-70\% & Avg. \\
\midrule
 & & & 67.82 & 61.09 & 68.95 & \cellcolor{yellow!40}60.64 & 73.55 & 67.74 & 70.65 & \cellcolor{yellow!40}69.22 & 72.06 & 58.15 & 65.36 & \cellcolor{yellow!40}64.83 \\
 \checkmark & & & 68.15 & 63.23 & 69.07 & \cellcolor{yellow!40}63.09 & 74.11 & 69.48 & 70.81 & \cellcolor{yellow!40}70.70 & 72.86 & 59.11 & 65.85 & \cellcolor{yellow!40}65.56 \\
 & \checkmark & & 68.95 & 64.68 & 70.93 & \cellcolor{yellow!40}64.40 & 74.31 & 70.65 & 71.81 & \cellcolor{yellow!40}72.17 & 72.90 & 58.99 & 65.77 & \cellcolor{yellow!40}65.49 \\
 & & \checkmark & 68.23 & 65.32 & 70.32 & \cellcolor{yellow!40}63.42 & 74.52 & 70.16 & 71.49 & \cellcolor{yellow!40}70.77 & 72.38 & 58.27 & 65.56 & \cellcolor{yellow!40}65.02 \\
 \checkmark & \checkmark & & 69.40 & 65.85 & 71.98 & \cellcolor{yellow!40}65.71 & 75.00 & 71.94 & 72.46 & \cellcolor{yellow!40}73.05 & 73.55 & 60.04 & 66.13 & \cellcolor{yellow!40}66.09 \\
 \checkmark & & \checkmark & 69.11 & 66.57 & 71.73 & \cellcolor{yellow!40}65.43 & 74.84 & 71.33 & 71.17 & \cellcolor{yellow!40}71.92 & 72.82 & 59.52 & 66.33 & \cellcolor{yellow!40}65.78 \\
 & \checkmark & \checkmark & 69.72 & 66.17 & 72.38 & \cellcolor{yellow!40}65.67 & 74.92 & 71.41 & 72.26 & \cellcolor{yellow!40}72.64 & 72.98 & 59.15 & 66.09 & \cellcolor{yellow!40}65.58 \\

\rowcolor{blue!20}
\checkmark & \checkmark & \checkmark
& \textbf{70.69} & \textbf{67.10} & \textbf{73.10}
& \textbf{66.73}
& \textbf{75.28} & \textbf{72.66} &  \textbf{72.42}
& \textbf{73.35}
&  \textbf{73.39} &  \textbf{59.88} &  \textbf{66.21}
& \textbf{66.13} \\

\bottomrule
\end{tabular}%
\end{adjustbox}
\end{table*}

\begin{figure*}[!t]
\centering
\includegraphics[width=0.95\linewidth]{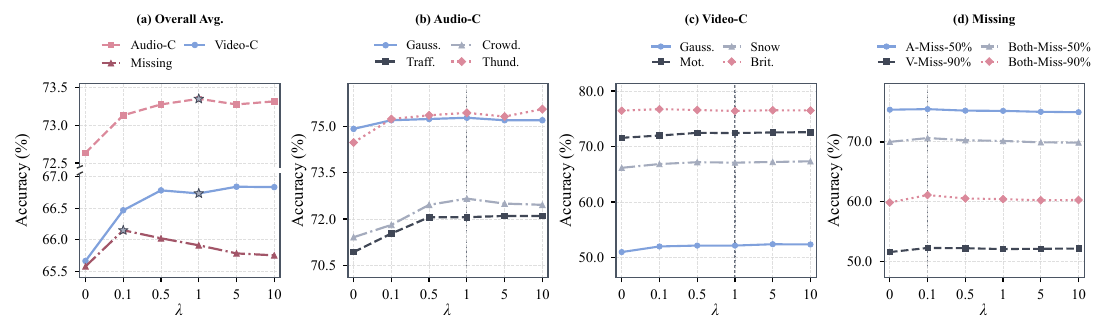}
\vspace{-0.7cm}
\caption{
Sensitivity analysis of $\lambda$ on Kinetics50. 
(a) Mean accuracy under audio corruption, video corruption, and missing-modality settings. 
(b–d) Accuracy under representative audio corruptions, video corruptions, and missing-modality conditions, respectively. 
}
  \label{fig:ab_lamb}
\end{figure*}

We first evaluate the overall performance of AdaPGC under different multimodal distribution shift scenarios. 
As shown in Tables~\ref{tab:ks50_video_c}-\ref{tab:audio_c}, AdaPGC achieves the best average performance under modality corruption scenarios, where video and audio modalities are independently evaluated under different corruption types.
Tables~\ref{tab:missing_ks50}-\ref{tab:missing_vgg} further report the results under modality missingness scenarios, where AdaPGC achieves consistent improvements across different missing patterns.

Under modality corruption scenarios, AdaPGC demonstrates consistent robustness against diverse types of visual and audio degradations. 
Across the corruption benchmarks, AdaPGC achieves competitive average accuracy in both video-corrupted and audio-corrupted settings on both datasets.
The improvements are consistently observed across different corruption types rather than being limited to specific corruption. 
For example, on Kinetics50-C with corrupted video inputs, AdaPGC improves the average accuracy from 65.9\% achieved by the strongest competing method to 66.7\%. 
For corrupted audio inputs on VGGSound-C, AdaPGC further improves the average accuracy from 36.5\% to 37.5\%, with consistent gains across various audio disturbances. 
These results indicate that modeling category-conditional representation distributions provides reliable calibration signals when modality information is affected by different types of corruption.

Under modality missingness scenarios, AdaPGC improves the adaptation performance when partial modality information becomes unavailable. 
Across different missing configurations and missing ratios, AdaPGC achieves the highest average accuracy on both Kinetics50 and VGGSound benchmarks.
The performance remains stable as the missing ratio increases, demonstrating the effectiveness of recovering latent fusion representations from the available modality information. 
For example, on Kinetics50 with the video modality missing in 90\% of samples, AdaPGC achieves 52.2\% accuracy, exceeding the best competing result by 1.3 percentage points.
Similarly, under the Both Missing setting with a 90\% missing ratio, AdaPGC improves accuracy from the best baseline result of 59.5\% to 61.1\%.
These results verify that the proposed cross-perspective representation recovery effectively preserves discriminative multimodal information even when complete multimodal inputs are unavailable.

\subsection{Ablation Study}

\begin{figure*}[!t]
\centering
\includegraphics[width=0.95\linewidth]{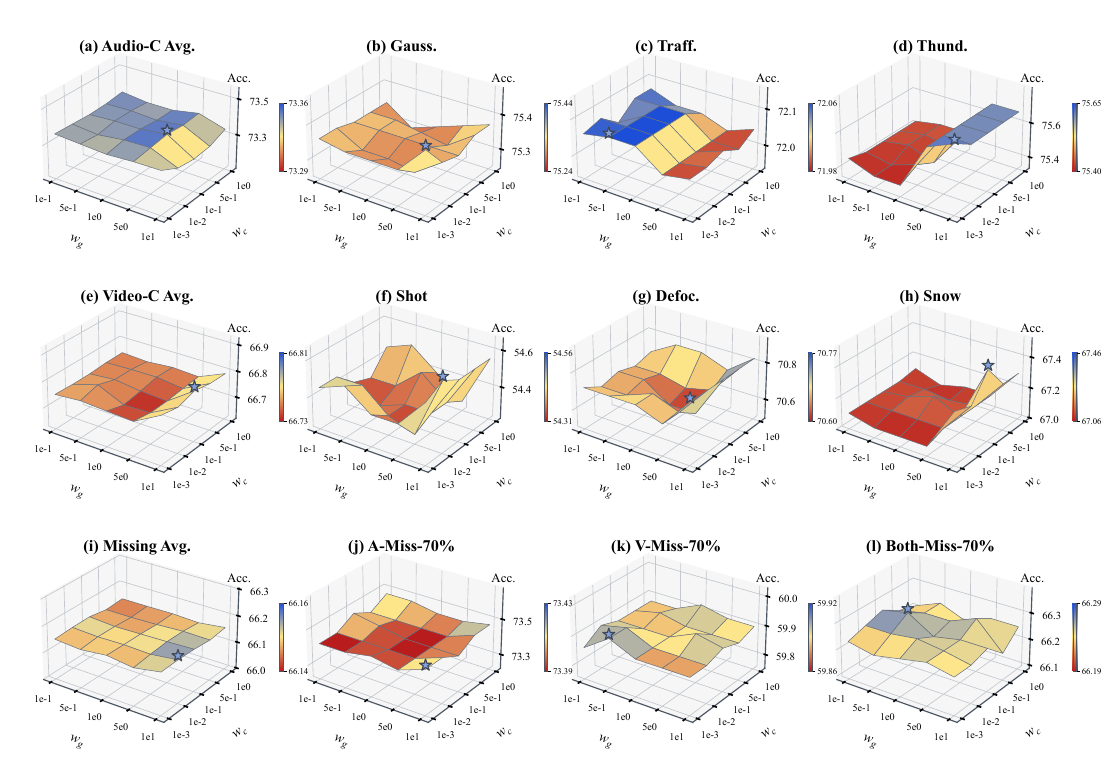}
\vspace{-0.7cm}
\caption{
Sensitivity of AdaPGC to $w_{\text{c}}$ and $w_{\text{g}}$ on Kinetics50. 
Rows correspond to audio corruption (a–d), video corruption (e–h), and missing-modality settings (i–l). 
The first column shows the accuracy averaged over all conditions in each setting, while the remaining columns present representative conditions. 
}
\label{fig:ab_wc_wg}
\end{figure*}

\begin{figure}[t]
  \centering
  \includegraphics[width=0.9\linewidth]{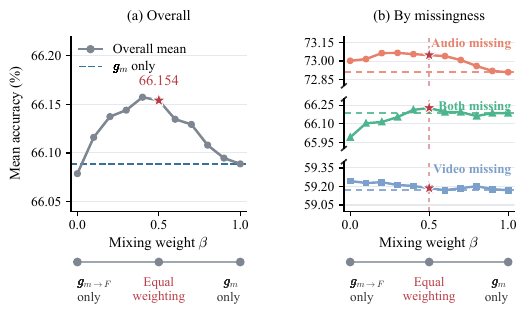}
  \vspace{-0.3cm}
  \caption{Sensitivity to the relative weighting of the modality-specific log-posterior and the recovery logits on Kinetics50. For incomplete samples, the Gaussian calibration logits are computed as
$\boldsymbol{r} =
(1-\beta)\boldsymbol{g}_{m\rightarrow F}
+\beta\boldsymbol{g}_{m}$.
(a) Mean accuracy over all 15 missing-modality conditions.
(b) Mean accuracy under each modality-missing setting, averaged over
missing rates of 50\%--90\%.
Horizontal dashed lines indicate the results using only
$\boldsymbol{g}_{m}$, while red stars mark the adopted equal weighting
($\beta=0.5$).
}
  \vspace{-0.3cm}
  \label{fig:ab-beta}
\end{figure}

We further conduct ablation studies to investigate the contribution of each component in AdaPGC. 
The results are summarized in Table~\ref{tab:components_ablation}. 
Specifically, we evaluate three components: the fused-logit calibration mechanism (FL; \eqref{eq:calibrated-logits}), which integrates the calibration logits with the original model predictions to refine the final output; the prediction alignment loss (PA; \eqref{eq:calibration-loss}), which uses the calibration prediction as supervision to guide test-time optimization; and the asymmetric rectification module (AR; \eqref{eq:onesided_infonce_instance}), which mitigates the impact of inconsistent modality reliability by rectifying the representation of the less reliable modality.
When all three components are included, AdaPGC achieves the highest average accuracy in each of the three evaluation settings; removing any component reduces the corresponding average accuracy.

FL consistently improves the baseline across all three settings, increasing the average accuracy by 2.45, 1.48, and 0.73 percentage points on Video-C, Audio-C, and Missing, respectively. Its contribution remains evident when PA is already enabled, where adding FL further improves the average accuracy from 64.40 to 65.71 on Video-C, from 72.17 to 73.05 on Audio-C, and from 65.49 to 66.09 under modality missingness. These results demonstrate that directly incorporating Gaussian calibration into the final prediction provides complementary information beyond adapting the model parameters alone.

PA yields the largest standalone improvement under modality corruption, raising the average accuracy by 3.76 percentage points on Video-C and 2.95 percentage points on Audio-C. This indicates that the probabilistic predictions derived from online Gaussian distributions provide effective supervision for adapting the source model to reliability changes at test time. 
Moreover, the consistent gains obtained by combining PA with FL show that Gaussian calibration serves two complementary roles: it can guide test-time model adaptation through PA while also directly calibrating the final prediction through FL.

AR provides a distinct contribution that is primarily concentrated on modality corruption. When used alone, it improves the average accuracy by 2.78 and 1.55 percentage points on Video-C and Audio-C, respectively, while producing only a marginal improvement under modality missingness. A similar pattern is observed when AR is added to FL and PA: the average accuracy increases from 65.71 to 66.73 on Video-C and from 73.05 to 73.35 on Audio-C, whereas the change under modality missingness is negligible. This behavior is consistent with the role of AR, which is specifically designed to rectify degraded modality representations rather than compensate for unavailable modalities. Its benefit is particularly evident under several corruption conditions, such as Snow, where AR alone improves the accuracy from 61.09 to 65.32.

Overall, FL, PA, and AR operate at complementary levels: FL directly calibrates the prediction, PA transfers probabilistic calibration into test-time model adaptation, and AR rectifies degraded modality representations. Combining all three components achieves the highest average accuracy on Video-C, Audio-C, and Missing, confirming their complementary contributions to robust adaptation under modality shifts.

\clearpage
\begin{figure*}[p]
  \centering
  \includegraphics[width=\linewidth]{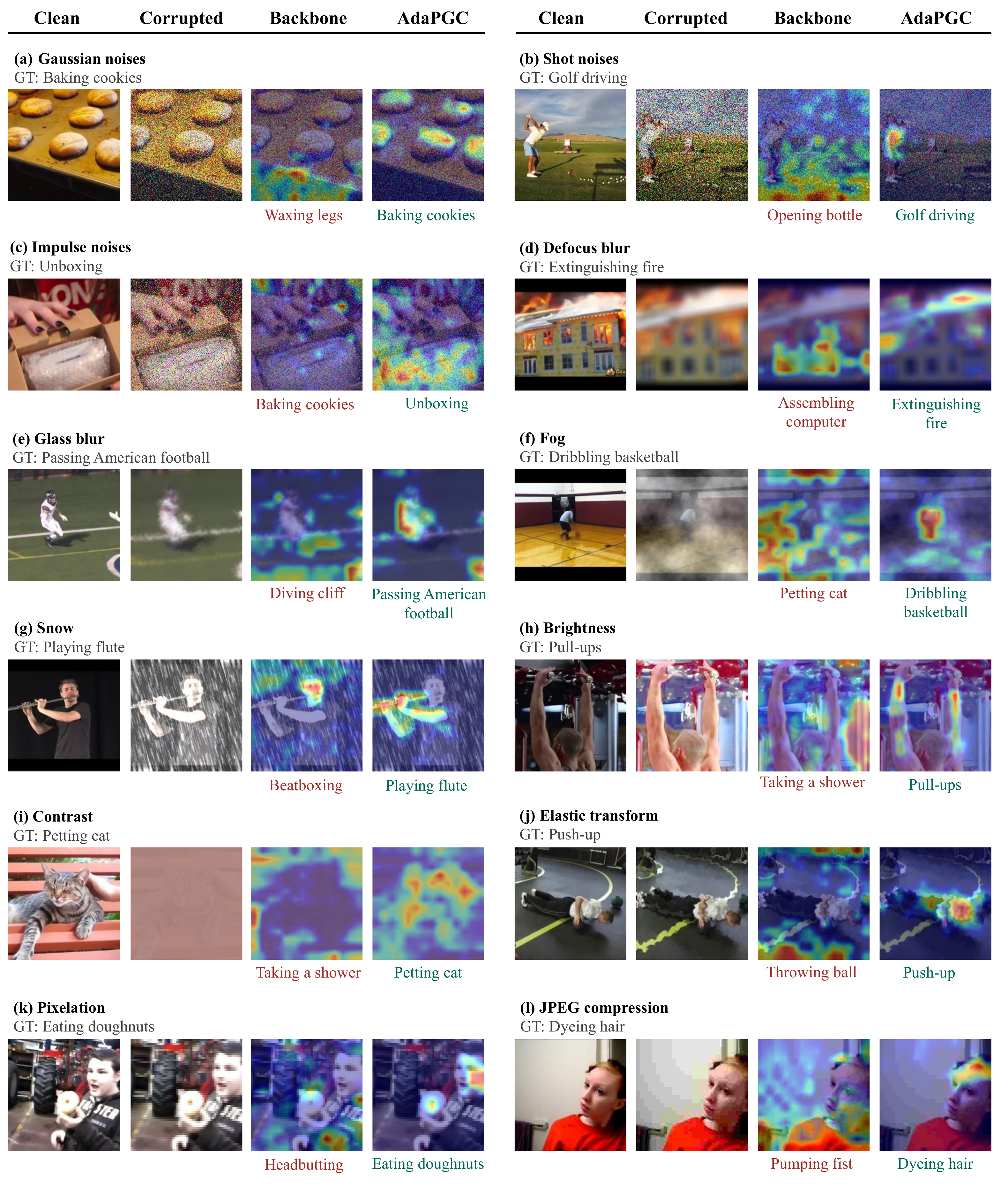}
\vspace{-0.7cm}
\caption{
Grad-CAM visualizations of the backbone and AdaPGC under twelve visual corruptions on Kinetics50-C. 
Each example shows the clean frame, corrupted input, and prediction-specific Grad-CAM overlays for the backbone and AdaPGC. 
GT denotes the ground-truth category; red and green labels indicate incorrect and correct predictions, respectively. 
In these selected correction cases, AdaPGC generally emphasizes action-related regions more clearly, whereas backbone responses are often dispersed or concentrated on the background. 
Maps are independently normalized for each branch’s predicted category, with warmer colors indicating larger relative positive contributions.
}
  \label{fig:exp_gradcam}
\end{figure*}
\clearpage

\subsection{Further Analysis}
\label{sec:exp-further}
\vskip -0.09in

\paragraph{Sensitivity to the Residual Calibration Weight}
We investigate the sensitivity of AdaPGC to the residual calibration weight $\lambda$, which scales the Gaussian calibration term added to the source-model logits.
As shown in Figure~\ref{fig:ab_lamb}, the effect of $\lambda$ differs between modality corruption and missingness.
Under modality corruption, increasing $\lambda$ beyond $1$ provides limited additional improvements, although higher accuracy can be obtained on Video-C with larger values.
Under modality missingness, a smaller calibration contribution is preferable, with $\lambda=0.1$ achieving the highest mean accuracy among the evaluated values.
These observations are consistent with the adopted settings of $\lambda=1$ for modality corruption and $\lambda=0.1$ for modality missingness.

\paragraph{Hyperparameter Sensitivity Analysis}

We further analyze the sensitivity of AdaPGC to the loss weights $w_g$ and $w_c$, which control Gaussian prediction alignment and reliability-aware representation rectification, respectively.
Figure~\ref{fig:ab_wc_wg} reports the results obtained by varying these two
hyperparameters over a predefined grid.
The optimal combinations of $w_{\text{c}}$ and $w_{\text{g}}$ vary across different corruption types and missing-modality conditions, as highlighted by the stars in each subfigure. 
However, when accuracy is averaged over all conditions within each evaluation setting, the performance surfaces become relatively smooth across the grid, indicating that AdaPGC maintains stable average performance without relying on precise tuning of these two weights.

\begin{figure*}[t]
  \centering
  \includegraphics[width=0.9\linewidth]{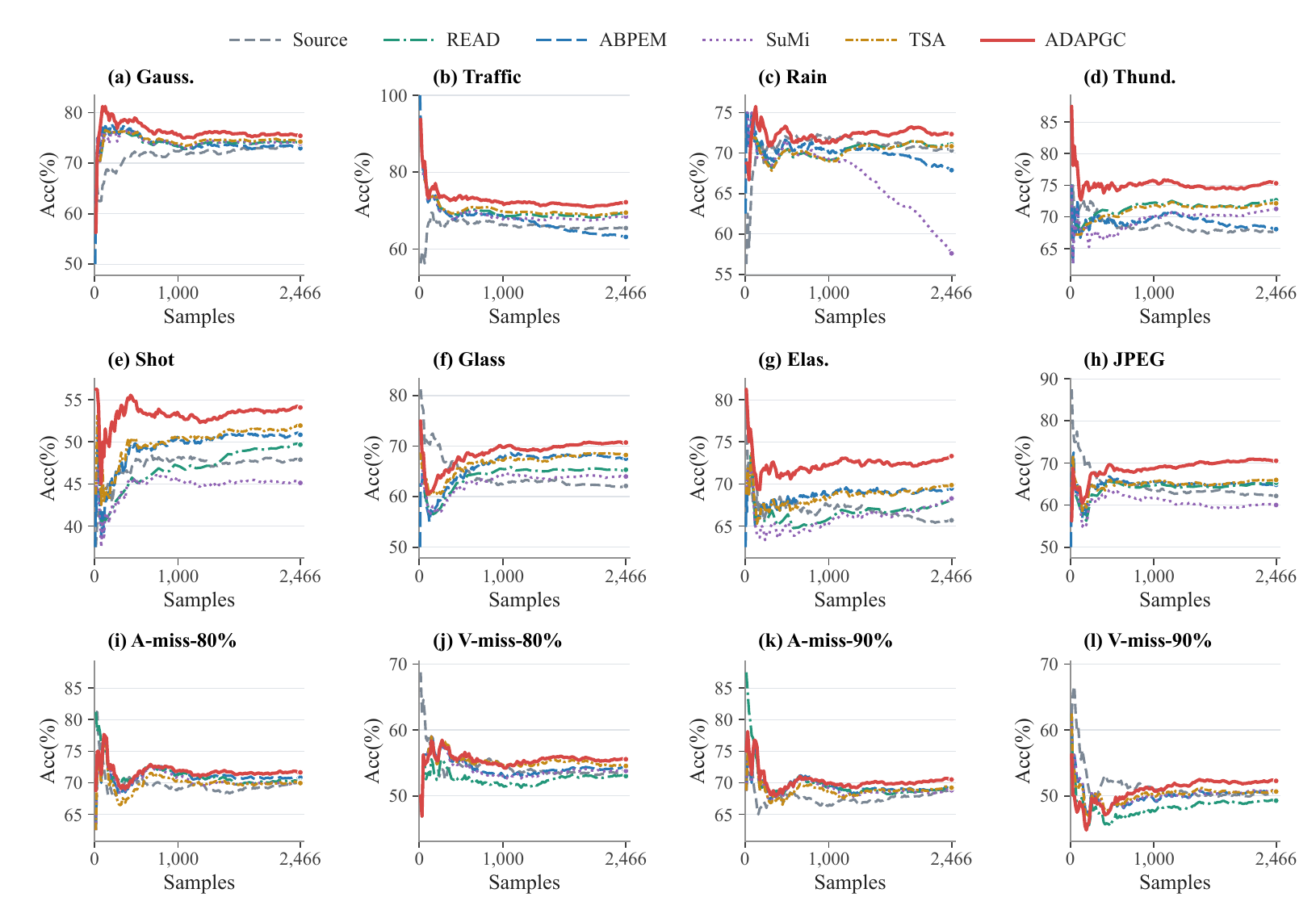}
\vspace{-0.2cm}
\caption{
Cumulative accuracy during test-time adaptation on Kinetics50. Rows correspond to audio corruption (a–d), video corruption (e–h), and missing-modality settings (i–l). 
A-miss and V-miss denote missing audio and visual modalities, respectively, with percentages indicating the missing rates. 
Each curve shows the accuracy over all test samples processed so far.
}
  \label{fig:exp_acc}
\end{figure*}

\begin{figure*}[t]
  \centering
  \includegraphics[width=0.9\linewidth]{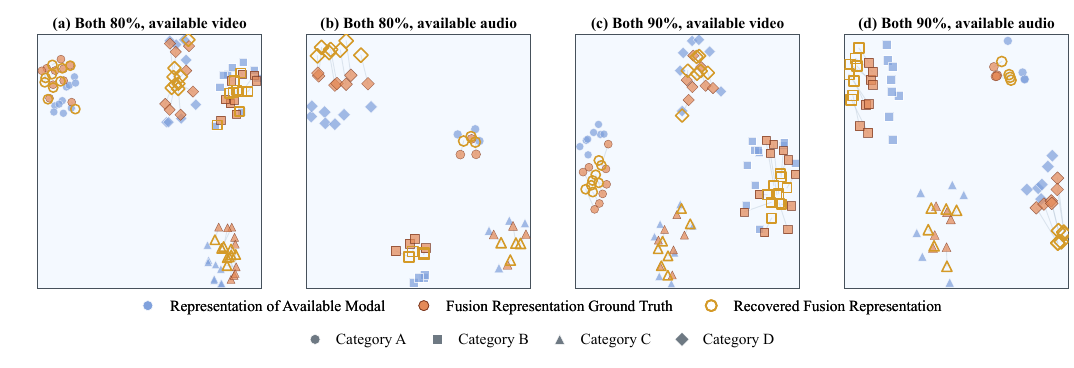}
\vspace{-0.7cm}
\caption{
Visualization of fusion representation recovery under multimodal missingness. 
Panels (a–b) and (c–d) show the 80\% and 90\% missing-both settings, respectively, using the available video or audio modality as the recovery source. 
Colors and fill styles distinguish representation types, marker shapes denote semantic categories, and gray lines connect each recovered representation to its paired clean ground truth. 
}
  \label{fig:exp_recovery}
\end{figure*}

\paragraph{Relative Weighting of Calibration Terms}
\label{sec:exp-weighting}
To examine the equal weighting adopted in \eqref{eq:equal-weighting}, we introduce a mixing coefficient $\beta$ and compute the Gaussian calibration logits for incomplete samples as $\boldsymbol{r}=(1-\beta)\boldsymbol{g}_{m\rightarrow F}+\beta\boldsymbol{g}_m$, where $\beta=0$ uses only the recovery logits and $\beta=1$ uses only the modality-specific log-posterior. 
As shown in Figure~\ref{fig:ab-beta}, the preferred weighting differs across modality-missing settings: Visual Missing favors the recovery logits, whereas Both Missing achieves its highest mean accuracy with equal weighting. 
These results suggest that the relative utility of the two calibration terms depends on the modality availability pattern. 
A paired $t$-test further supports a statistically significant improvement in overall mean accuracy over using only $\boldsymbol{g}_m$ at the 95\% confidence level ($p=0.0156$).
The adopted setting $\beta=0.5$ balances these setting-dependent preferences and achieves near-optimal overall accuracy, supporting its use as a common configuration across different modality-missing settings.

\begin{figure}[t]
  \centering
  \includegraphics[width=0.9\linewidth]{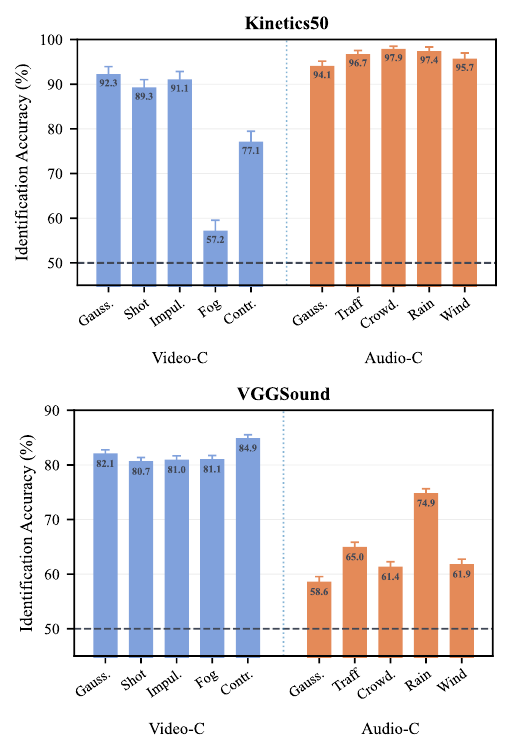}
  \vspace{-0.5cm}
  \caption{
Unreliable modality identification accuracy on Kinetics50-C and VGGSound-C under video and audio corruptions. The dashed horizontal line indicates the 50\% chance level, and error bars represent 95\% confidence intervals.
  }
  \label{fig:exp_rel_ide}
\end{figure}

\paragraph{Case Study }
We use Grad-CAM to examine the spatial evidence associated with the predictions of the backbone and AdaPGC under modality reliability shifts. 
As shown in Figure~\ref{fig:exp_gradcam}, twelve representative samples from Kinetics50-C are selected across different visual corruptions, covering cases where AdaPGC corrects an erroneous backbone prediction. 
The visualizations reveal differences in the regions contributing to the respective predictions. 
In the illustrated examples, the backbone exhibits responses extending beyond action-related regions, whereas AdaPGC shows comparatively stronger emphasis on regions relevant to the target action. 
For example, under shot noise and fog, AdaPGC emphasizes the golfer in Figure~\ref{fig:exp_gradcam} (b) and the basketball player in Figure~\ref{fig:exp_gradcam} (f), whereas the backbone responds more strongly to surrounding regions. 
In Figure~\ref{fig:exp_gradcam} (g), the backbone misclassifies playing flute as beatboxing and primarily highlights the face, while AdaPGC emphasizes the flute and arms, providing spatial evidence more consistent with the correct action. 
Similar patterns are observed under elastic transformation and JPEG compression, where AdaPGC highlights the person performing the push-up and the hair region, respectively. 
These selected cases illustrate how successful prediction calibration is accompanied by spatial attribution patterns that better correspond to recognizable action-related content, complementing the quantitative evaluation of robustness to visual corruption. 
Each map explains the corresponding branch’s predicted category and is independently normalized, so the comparison concerns qualitative spatial patterns rather than absolute attribution magnitudes.

\paragraph{Adaptation Performance Over the Test Stream}
To examine how adaptation performance evolves as test samples arrive, we compare the cumulative accuracy of AdaPGC with the source model and representative multimodal TTA methods. 
Figure~\ref{fig:exp_acc} presents results under selected audio corruptions, video corruptions, and missing-modality conditions. 
Despite initial fluctuations, AdaPGC achieves the highest final cumulative accuracy across all displayed conditions. 
Under modality corruption, AdaPGC maintains a clear advantage in most cases, particularly for thunder noise, shot noise, and JPEG compression. 
Some competing methods exhibit declining accuracy as adaptation proceeds, such as SuMi under rain noise, whereas AdaPGC maintains relatively stable performance. 
Moreover, under glass blur, elastic transformation, and JPEG compression, AdaPGC exhibits an overall upward trend in cumulative accuracy after the initial fluctuations, indicating that its predictions improve as adaptation proceeds. 
Under modality missingness, AdaPGC also achieves consistent improvements at both 80\% and 90\% missing rates. 
These results complement the aggregate comparisons and demonstrate the effectiveness of AdaPGC throughout adaptation under corrupted and incomplete multimodal inputs.

\paragraph{Recovered Fusion Representation Analysis}
We further analyze the recovered fusion representations under modality missingness scenarios. 
Since the proposed recovery process estimates category-conditional fusion representations 
$\hat{\boldsymbol{z}}_{F|m,c,i}$ for each possible category, we aggregate them using the modality-specific posterior probabilities for visualization:
$
\hat{\boldsymbol{z}}_{F,i}
=
\sum_{c=1}^{C}
\mathrm{Softmax}(\boldsymbol{g}_m(\boldsymbol{z}_{m,i}))_c
\hat{\boldsymbol{z}}_{F|m,c,i}.
$
The visualization compares the aggregated recovered representations with the ground-truth fusion representations obtained from complete multimodal inputs. 
As shown in Figure~\ref{fig:exp_recovery}, the recovered representations remain close to their corresponding ground-truth fusion representations and preserve the category-wise clustering structures.
Moreover, this alignment is maintained under different missing ratios, indicating that the proposed category-conditional cross-perspective recovery captures meaningful statistical relationships between available modality representations and fusion representations.

\paragraph{Unreliable Modality Identification}

We evaluate the effectiveness of the proposed distribution discrepancy criterion in identifying unreliable modalities under different corruption scenarios. 
As shown in Figure~\ref{fig:exp_rel_ide}, the identification accuracy remains consistently above the 50\% chance level across different corruption types and datasets, demonstrating that the discrepancy between modality-specific and multimodal predictions provides informative signals for reliability assessment. 
The variations among corruption types indicate that different corruptions induce different levels of prediction discrepancy, while the overall effectiveness of the criterion remains stable. 
These results validate the reliability identification strategy used in the asymmetric rectification module.

\section{Conclusion}

This work presents AdaPGC, a probabilistic multimodal test-time adaptation framework for distribution shifts involving variations in both modality reliability and availability.
AdaPGC continuously models category-conditional distributions over modality-specific and multimodal fusion representations, using the resulting probabilistic predictions to calibrate source model outputs throughout the target stream.
Beyond prediction calibration, AdaPGC exploits cross-perspective distributional information to rectify degraded modality representations and further models the joint distributions between modality-specific and fusion representations to recover unavailable fusion information under modality missingness.
Extensive experiments across diverse corruption and missingness conditions demonstrate that AdaPGC consistently improves adaptation robustness over existing approaches, validating the effectiveness of probabilistic distribution modeling for multimodal test-time adaptation under complex modality shifts.

\bibliographystyle{IEEEtran}
\bibliography{main}

\vskip -0.5in
\begin{IEEEbiography}[{\includegraphics[width=1in,height=1.25in,clip,keepaspectratio]{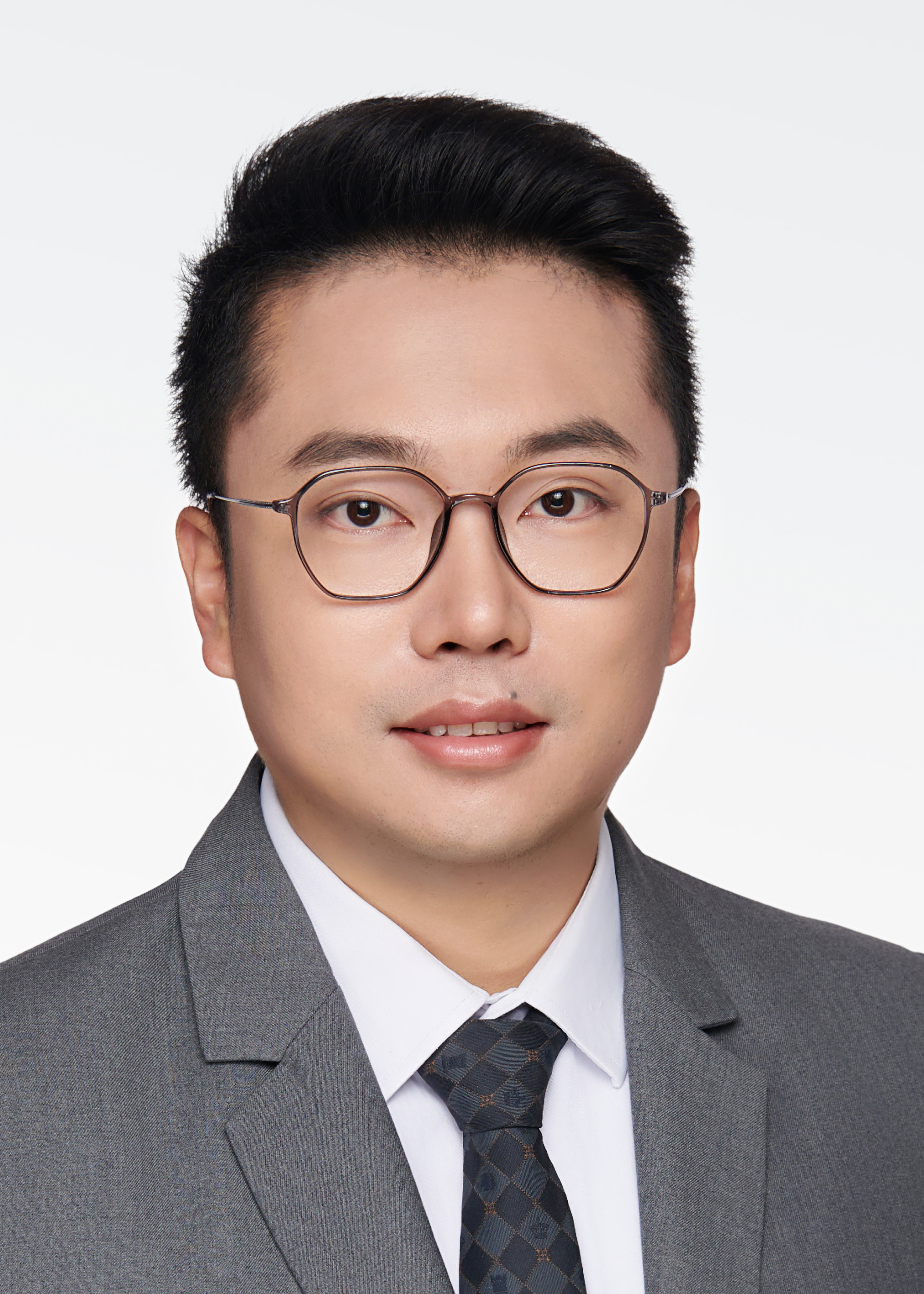}}]{Jiangmeng Li}
received the BS degree from Xiamen University in 2016, the MS degree from New York University in 2018, and the Ph.D. degree from University of Chinese Academy of Sciences in 2023. He is currently an Associate Professor (tenure-track) at the Institute of Software, Chinese Academy of Sciences. His research interests include multi-modal mearning, self-supervised learning, and graph learning. He has published more than 70 papers in journals and conferences such as IEEE TPAMI, IEEE TKDE, IJCV, ICML, NeurIPS, and etc. 7 papers recognized as Spotlight Papers at ICML, NeurIPS, KDD, and AAAI. In academic service, he serves as Area Chair for ICLR 2027, Session Chair for AAAI 2026, Senior Program Committee for AAAI, reviewer or Program Committee for prestigious journals and conferences such as IEEE TPAMI, IJCV, NeurIPS, ICML, and ICLR. He also serves as Editorial Board Member of the international SCI journal Mathematics and International Standard Editor to the International Telecommunication Union (ITU).
\end{IEEEbiography}

\vskip -0.5in
\begin{IEEEbiography}[{\includegraphics[width=1in,height=1.25in,clip,keepaspectratio]{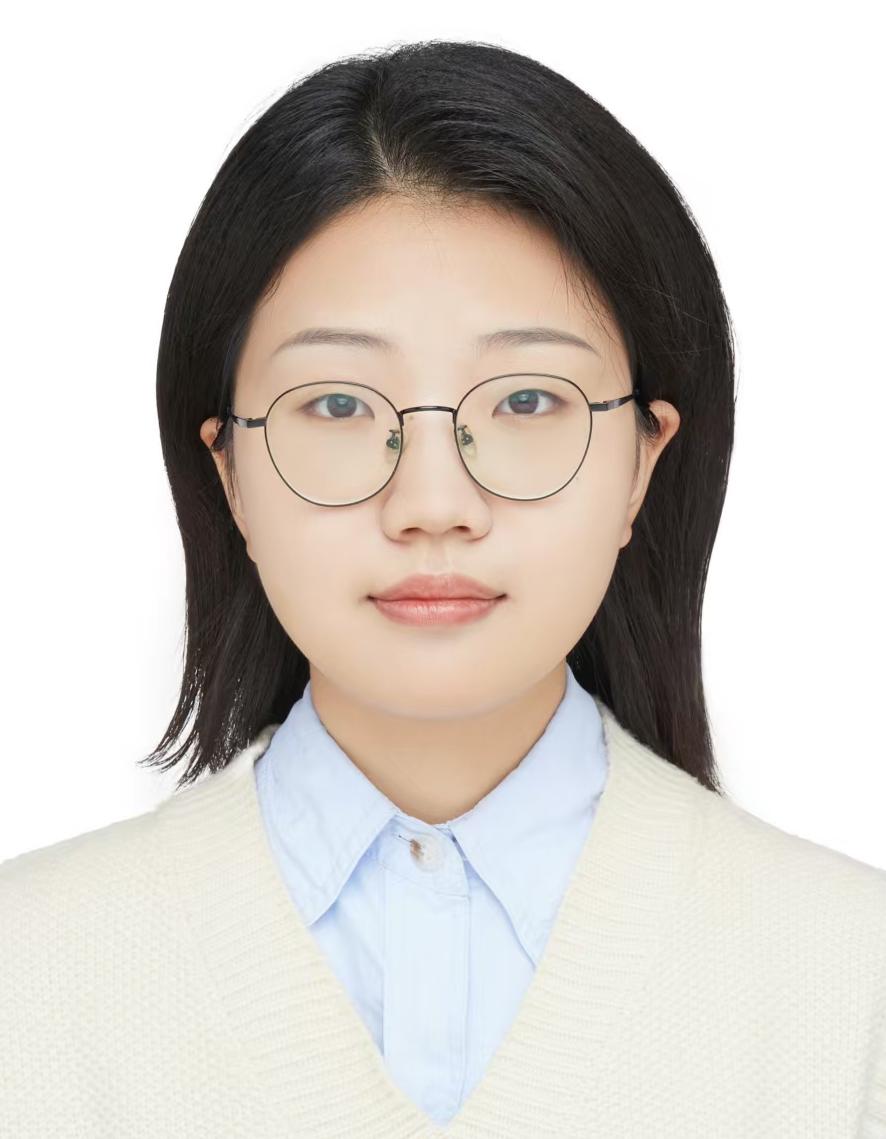}}]{Jinglin Xu}
received the B.S. degree from Beihang University, Beijing, China, in 2023. She is currently pursuing the Ph.D. degree at the University of Chinese Academy of Sciences and the Institute of Software, Chinese Academy of Sciences, Beijing, China. Her research interests include multimodal representation learning, self-supervised learning, and test-time adaptation.
\end{IEEEbiography}

\vskip -0.7in
\begin{IEEEbiography}[{\includegraphics[width=1in,height=1.25in,clip,keepaspectratio]{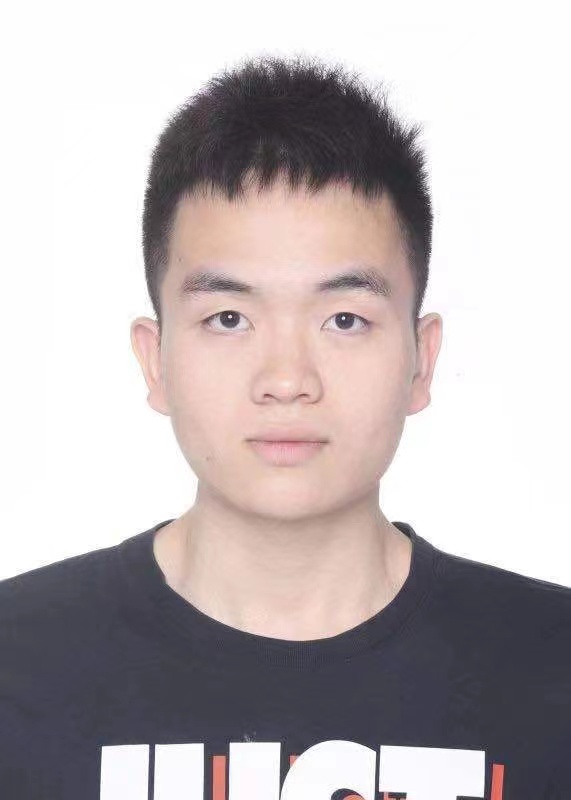}}]{Yi Li} received the B.Sc degree in the School of Fuzhou University, Fuzhou, China. He is currently working toward the Ph.D degree in University of Chinese Academy of Sciences, Beijing, China. His research interests lie in the domains of multimodal learning and causality, with a particular focus on vision-language representation learning and causal representation learning.
\end{IEEEbiography}

\vskip -0.5in
\begin{IEEEbiography}
[{\includegraphics[width=1in,height=1.25in,clip,keepaspectratio]{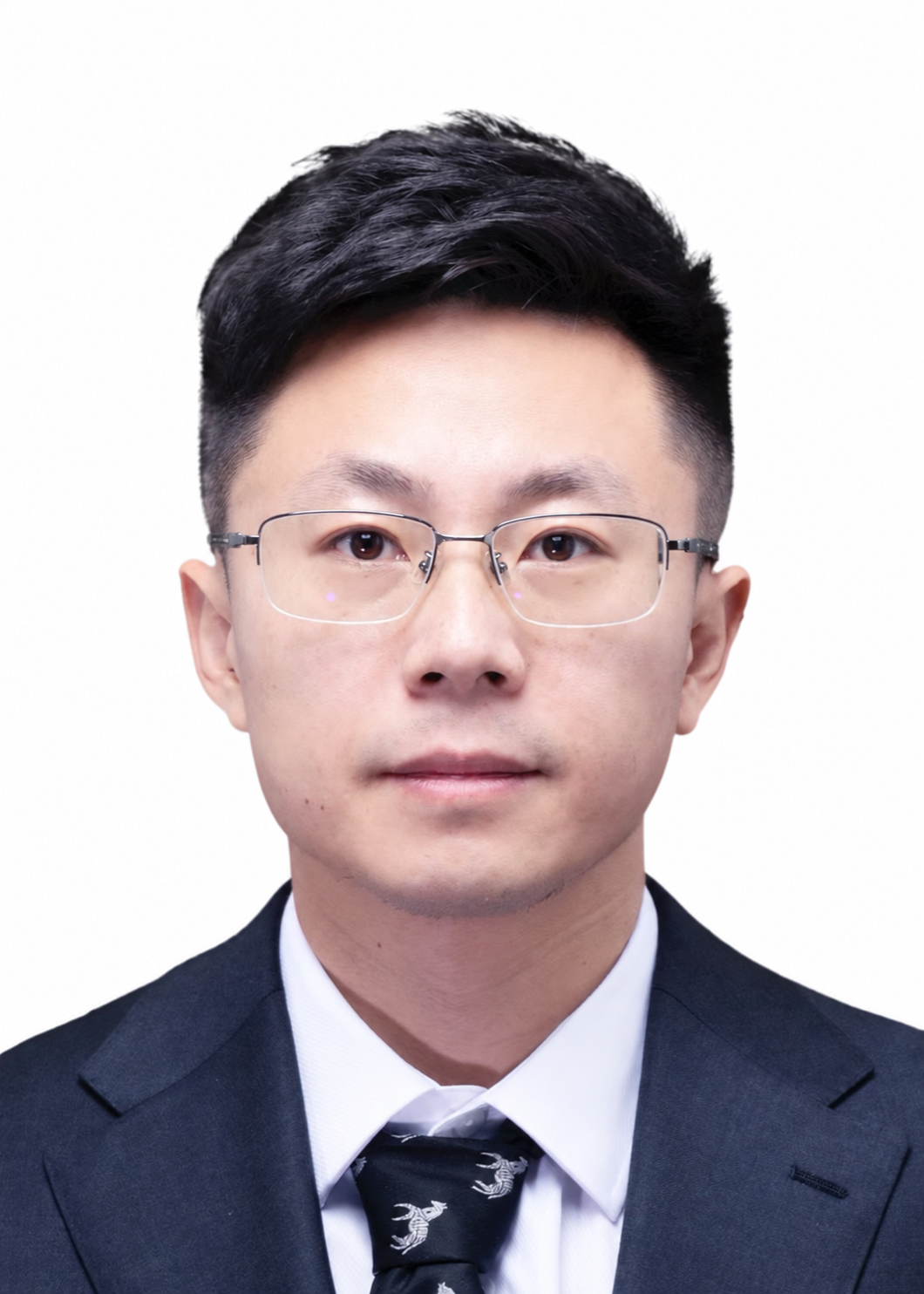}}]{Yumeng Liu} 
received the Ph.D. degree in computer science from the University of Chinese Academy of Sciences and the M.S. degree in computer science from the University of Southern California. He is currently a Senior Engineer at the Institute of Software, Chinese Academy of Sciences. His main research fields are cloud computing, database systems, and artificial intelligence. He has published more than 21 papers in international conferences and journals.
\end{IEEEbiography}

\vskip -0.7in
\begin{IEEEbiography}[{\includegraphics[width=1in,height=1.25in,clip,keepaspectratio]{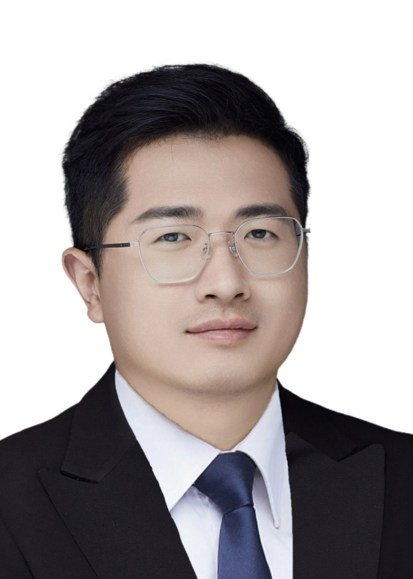}}]{Jiahuan Zhou}
received his B.E. degree from the Department of Automation, Tsinghua University, in 2013, and his Ph.D. degree from the Department of Electrical Engineering and Computer Science, Northwestern University, in 2018. He is currently a Tenure-Track Assistant Professor at the Wangxuan Institute of Computer Technology, Peking University. His research interests lie at the intersection of computer vision, multimodal learning, and AI4Science. He has published more than 90 papers in leading international journals and conferences, including Nature Synthesis, Nature Communications, IEEE TPAMI, IJCV, IEEE TIP, CVPR, ICCV, ECCV, ICML, NeurIPS, ICLR, and others. He has served as a Workshop Chair for ICCV 2027, an Area Chair for CVPR, ECCV, ICML, NeurIPS, ICLR, BMVC, ICME, and ICPR, and as a Senior Program Committee Member for AAAI. He also serves as an Associate Editor for Machine Vision and Applications (MVA), and as a Youth Associate Editor for National Materials. His honors and recognitions include the National Overseas Excellent Young Scholars, the Xiaomi-Boya Young Scholar Award, and the Yang Fuqing–Wang Yangyuan Academician Faculty Award Excellence Prize.

\end{IEEEbiography}

\vskip -0.5in
\begin{IEEEbiography}[{\includegraphics[width=1in,height=1.25in,clip,keepaspectratio]{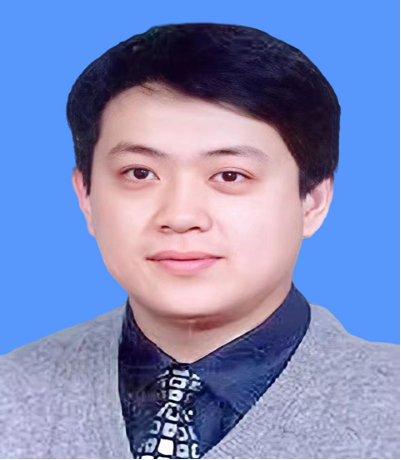}}]{Fanjiang Xu}
received his B.Sc. and M.Sc. degrees in computer science both from National University of Defense Technology and his Ph.D. degree in computer science from Huazhong University of Science and Technology. He is currently a Professor at the Institute of Software, Chinese Academy of Sciences. His research interests include intelligent information processing and integration.
\end{IEEEbiography}

\vskip -0.5in
\begin{IEEEbiography}[{\includegraphics[width=1in,height=1.25in,clip,keepaspectratio]{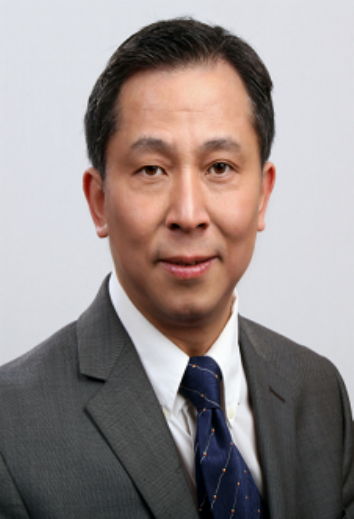}}]{Hui Xiong}
received his Ph.D. in Computer Science from the University of Minnesota - Twin Cities, USA, in 2005, the B.E. degree in Automation from the University of Science and Technology of China (USTC), Hefei, China, and the M.S. degree in Computer Science from the National University of Singapore (NUS), Singapore. He is a chair professor at the Hong Kong University of Science and Technology (Guangzhou). He is also a Distinguished Professor at Rutgers, the State University of New Jersey, where he received the 2018 Ram Charan Management Practice Award as the Grand Prix winner from the Harvard Business Review, RBS Dean's Research Professorship (2016), two-year early promotion/tenure (2009), the Rutgers University Board of Trustees Research Fellowship for Scholarly Excellence (2009), the ICDM-2011 Best Research Paper Award (2011), the Junior Faculty Teaching Excellence Award (2007), Dean's Award for Meritorious Research (2010, 2011, 2013, 2015) at Rutgers Business School, the 2017 IEEE ICDM Outstanding Service Award (2017), and the AAAI-2021 Best Paper Award (2021). Dr. Xiong is also a Distinguished Guest Professor (Grand Master Chair Professor) at the University of Science and Technology of China (USTC). For his outstanding contributions to data mining and mobile computing, he was elected an ACM Distinguished Scientist in 2014, an IEEE Fellow and an AAAS Fellow in 2020. His general area of research is data and knowledge engineering, with a focus on developing effective and efficient data analysis techniques for emerging data intensive applications. He currently serves as a co-Editor-in-Chief of Encyclopedia of GIS (Springer) and an Associate Editor of IEEE Transactions on Data and Knowledge Engineering (TKDE), IEEE Transactions on Big Data (TBD), ACM Transactions on Knowledge Discovery from Data (TKDD) and ACM Transactions on Management Information Systems (TMIS). He has served regularly on the organization and program committees of numerous conferences, including as a Program Co-Chair of the Industrial and Government Track for the 18th ACM SIGKDD International Conference on Knowledge Discovery and Data Mining (KDD), a Program Co-Chair for the IEEE 2013 International Conference on Data Mining (ICDM), a General Co-Chair for the IEEE 2015 International Conference on Data Mining (ICDM), and a Program Co-Chair of the Research Track for the 24th ACM SIGKDD International Conference on Knowledge Discovery and Data Mining (KDD2018).
\end{IEEEbiography}

\vskip -0.5in

\begin{IEEEbiography}
[{\includegraphics[width=1.0in,height=1.25in,clip,keepaspectratio]{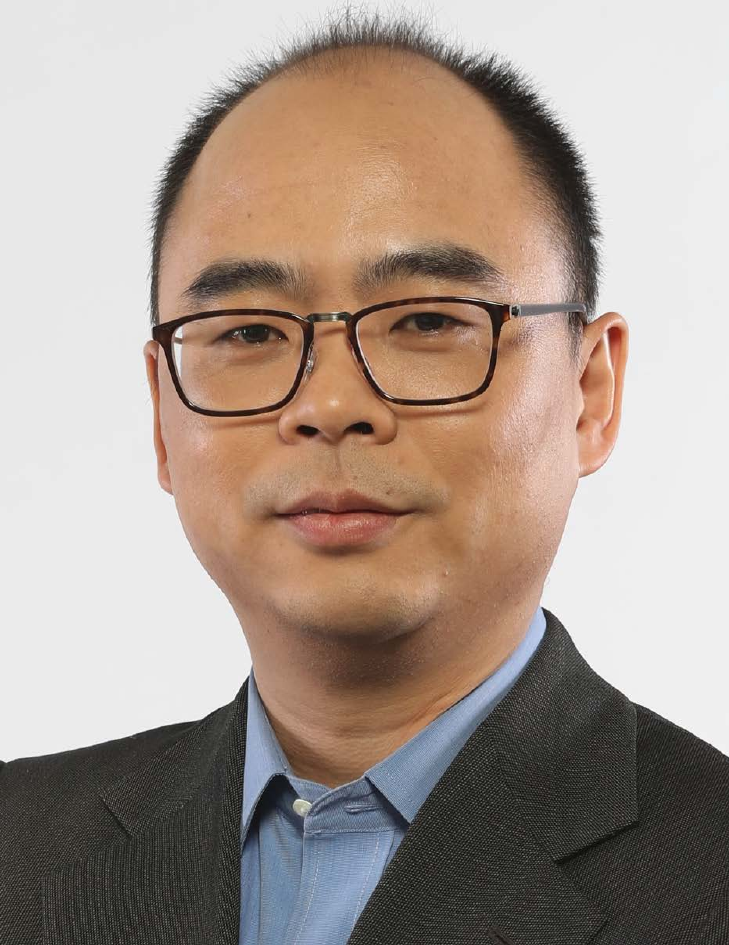}}]{Gang Hua} (Fellow, IEEE) received the B.S. and M.S. degrees in automatic control engineering from Xi’an Jiaotong University, Xi’an, China, in 1999 and 2002, respectively, and the Ph.D. degree in electrical engineering and computer science with Northwestern University, Evanston, IL, USA, in 2006. He is currently a Director of Applied Science at Amazon Alexa AI. He is an Associate Editor for TPAMI, IJCV and MVA. He is the General Chair of ICCV’2027, the Program Chair of CVPR’2019\&2022. He is the author of more than 280 peer-reviewed publications in prestigious international journals and conferences. He holds 37 U.S. patents and has 15 more U.S. patents pending. He was the recipient of the 2015 IAPR Young Biometrics Investigator Award. He is an IEEE Fellow, IAPR Fellow and an ACM Distinguished Scientist.
\end{IEEEbiography}

\clearpage

\appendices
\section{Gaussian Fusion Estimation and Prediction Aggregation under Modality Missingness}
\label{appendix:recovery-derivation}
We provide the probabilistic motivation and derivation for the posterior-weighted recovery logits in Eq.~\eqref{eq:recovery-logit}.
Without loss of generality, we consider the case where modality $m_2$ is unavailable and only the representation $\boldsymbol{z}_{m_1}$ of modality $m_1$ is observable.
The corresponding fusion representation $\boldsymbol{z}_{F}$ cannot be directly obtained.
For clarity, we omit the batch superscript $t$ and sample index $i$ throughout this derivation.

Given only the observable modality-specific representation $\boldsymbol{z}_{m_1}$, its category posterior can be directly obtained from the corresponding Gaussian mixture as

\begin{equation}
\label{eq:appendix-modality-posterior}
p_{m_1}(c\mid\boldsymbol{z}_{m_1})
=
\frac{
\pi_{m_1,c}
\mathcal{N}
\left(
\boldsymbol{z}_{m_1};
\boldsymbol{\mu}_{m_1,c},
\boldsymbol{\Sigma}_{m_1,c}
\right)
}{
\sum_{c'=1}^{C}
\pi_{m_1,c'}
\mathcal{N}
\left(
\boldsymbol{z}_{m_1};
\boldsymbol{\mu}_{m_1,c'},
\boldsymbol{\Sigma}_{m_1,c'}
\right)
}.
\end{equation}

This posterior characterizes category discrimination solely in the modality-specific representation space.
However, the fusion representation $\boldsymbol{z}_{F}$ normally incorporates multimodal information that is unavailable from $\boldsymbol{z}_{m_1}$ alone.
To further exploit the category distribution learned in the fusion representation space, we treat $\boldsymbol{z}_{F}$ as a latent variable and marginalize over its possible values.
The resulting recovery prediction is

\begin{equation}
\label{eq:appendix-marginalization}
p_{m_1\rightarrow F}
\left(
c
\mid
\boldsymbol{z}_{m_1}
\right)
=
\int
p
\left(
c
\mid
\boldsymbol{z}_{m_1},
\boldsymbol{z}_{F}
\right)
p
\left(
\boldsymbol{z}_{F}
\mid
\boldsymbol{z}_{m_1}
\right)
d\boldsymbol{z}_{F}.
\end{equation}

We assume that, conditioned on the fusion representation, the observable modality does not provide additional category information,
$
c
\perp
\boldsymbol{z}_{m_1}
\mid
\boldsymbol{z}_{F},
$
such that
$
p
\left(
c
\mid
\boldsymbol{z}_{m_1},
\boldsymbol{z}_{F}
\right)
=
p
\left(
c
\mid
\boldsymbol{z}_{F}
\right).
$
Equation~\eqref{eq:appendix-marginalization} therefore becomes
\begin{equation}
\label{eq:appendix-marginalization-simplified}
p_{m_1\rightarrow F}
\left(
c
\mid
\boldsymbol{z}_{m_1}
\right)
=
\int
p_{F}
\left(
c
\mid
\boldsymbol{z}_{F}
\right)
p
\left(
\boldsymbol{z}_{F}
\mid
\boldsymbol{z}_{m_1}
\right)
d\boldsymbol{z}_{F},
\end{equation}
where $p(\boldsymbol{z}_{F}\mid\boldsymbol{z}_{m_1})$ describes the possible fusion representations associated with the observed modality-specific representation.

We introduce the latent category variable $k$ and decompose this distribution according to the law of total probability:
\begin{equation}
\label{eq:appendix-latent-category}
p
\left(
\boldsymbol{z}_{F}
\mid
\boldsymbol{z}_{m_1}
\right)
=
\sum_{k=1}^{C}
p
\left(
\boldsymbol{z}_{F}
\mid
\boldsymbol{z}_{m_1},
k
\right)
p_{m_1}
\left(
k
\mid
\boldsymbol{z}_{m_1}
\right).
\end{equation}
To obtain a tractable form for
$p(\boldsymbol{z}_{F}\mid\boldsymbol{z}_{m_1},k)$,
we further model the modality-specific and fusion representations as jointly Gaussian within each category:
\begin{equation}
\begin{bmatrix}
\boldsymbol{z}_{m_1}
\\
\boldsymbol{z}_{F}
\end{bmatrix}
\Bigg|\, k
\sim
\mathcal{N}
\left(
\begin{bmatrix}
\boldsymbol{\mu}_{m_1,k}
\\
\boldsymbol{\mu}_{F,k}
\end{bmatrix},
\begin{bmatrix}
\boldsymbol{\Sigma}_{m_1,k}
&
\boldsymbol{\Sigma}_{m_1F,k}
\\
\boldsymbol{\Sigma}_{Fm_1,k}
&
\boldsymbol{\Sigma}_{F,k}
\end{bmatrix}
\right).
\end{equation}
Under this category-conditional joint Gaussian assumption, the conditional distribution of the latent fusion representation is also Gaussian:
\begin{equation}
\label{eq:appendix-conditional-gaussian}
p
\left(
\boldsymbol{z}_{F}
\mid
\boldsymbol{z}_{m_1},
k
\right)
=
\mathcal{N}
\left(
\boldsymbol{z}_{F};
\hat{\boldsymbol{z}}_{F\mid m_1,k},
\boldsymbol{\Sigma}_{F\mid m_1,k}
\right),
\end{equation}
with conditional mean
\begin{equation}
\label{eq:appendix-conditional-mean}
\hat{\boldsymbol{z}}_{F\mid m_1,k}
=
\boldsymbol{\mu}_{F,k}
+
\boldsymbol{\Sigma}_{Fm_1,k}
\boldsymbol{\Sigma}_{m_1,k}^{-1}
\left(
\boldsymbol{z}_{m_1}
-
\boldsymbol{\mu}_{m_1,k}
\right).
\end{equation}
Consequently,
\begin{equation}
\label{eq:appendix-fusion-mixture}
\footnotesize
p
\left(
\boldsymbol{z}_{F}
\mid
\boldsymbol{z}_{m_1}
\right)
=
\sum_{k=1}^{C}
p_{m_1}
\left(
k
\mid
\boldsymbol{z}_{m_1}
\right)
\mathcal{N}
\left(
\boldsymbol{z}_{F};
\hat{\boldsymbol{z}}_{F\mid m_1,k},
\boldsymbol{\Sigma}_{F\mid m_1,k}
\right).
\end{equation}

Substituting Eq.~\eqref{eq:appendix-fusion-mixture} into Eq.~\eqref{eq:appendix-marginalization-simplified} yields
\begin{equation}
\footnotesize
\begin{aligned}
&p_{m_1\rightarrow F}(c|z_{m_1})
\\
&=
\int
p_F(c|z_F)
\sum_{k=1}^{C}
p_{m_1}(k|z_{m_1})
\mathcal{N}
(z_F;\hat{z}_{F|m_1,k},
\Sigma_{F|m_1,k})
dz_F
\\
&=
\sum_{k=1}^{C}
p_{m_1}(k|z_{m_1})
\int
p_F(c|z_F)
\mathcal{N}
(z_F;\hat{z}_{F|m_1,k},
\Sigma_{F|m_1,k})
dz_F
\\
&=
\sum_{k=1}^{C}
p_{m_1}(k|z_{m_1})
\mathbb{E}_{
\boldsymbol{z}_{F}
\sim
\mathcal{N}
\left(
\hat{\boldsymbol{z}}_{F\mid m_1,k},
\boldsymbol{\Sigma}_{F\mid m_1,k}
\right)
}
\left[
p_{F}
\left(
c
\mid
\boldsymbol{z}_{F}
\right)
\right].
\end{aligned}
\end{equation}
When the conditional distribution is sufficiently concentrated around its mean, the latent fusion representation can be approximated by its conditional mean,
$
\boldsymbol{z}_{F}
\approx
\hat{\boldsymbol{z}}_{F\mid m_1,k},
$
which leads to the plug-in approximation
\begin{equation}
\footnotesize
\mathbb{E}_{
\boldsymbol{z}_{F}
\sim
\mathcal{N}
\left(
\hat{\boldsymbol{z}}_{F\mid m_1,k},
\boldsymbol{\Sigma}_{F\mid m_1,k}
\right)
}
\left[
p_{F}
\left(
c
\mid
\boldsymbol{z}_{F}
\right)
\right]
\approx
p_{F}
\left(
c
\mid
\hat{\boldsymbol{z}}_{F\mid m_1,k}
\right).
\end{equation}
Therefore, the recovery prediction can be approximated as
\begin{equation}
\footnotesize
\label{eq:appendix-recovery-posterior}
\begin{aligned}
p_{m_1\rightarrow F}
\left(
c
\mid
\boldsymbol{z}_{m_1}
\right)
\approx
\sum_{k=1}^{C}
p_{m_1}
\left(
k
\mid
\boldsymbol{z}_{m_1}
\right)
p_{F}
\left(
c
\mid
\hat{\boldsymbol{z}}_{F\mid m_1,k}
\right)
\end{aligned}
\end{equation}

Equation~\eqref{eq:appendix-recovery-posterior} shows that the prediction associated with the unavailable fusion representation can be approximated by a posterior-weighted combination of category-conditional recovery branches.
Specifically, $p_{m_1}(k\mid\boldsymbol{z}_{m_1})$ measures the plausibility of category $k$ given the observed modality, while 
$p_{F}(c\mid\hat{\boldsymbol{z}}_{F\mid m_1,k})$ 
gives the posterior probability of output category $c$ for the corresponding recovered representation under the Gaussian model in the fusion space.

Using the log-posterior defined in the main text, these posterior probabilities can be written as
\begin{equation}
\footnotesize
p_{m_1}
\left(
k
\mid
\boldsymbol{z}_{m_1}
\right)
=
\frac{
\exp
\left(
g_{m_1,k}
\left(
\boldsymbol{z}_{m_1}
\right)
\right)
}{
\sum_{d=1}^{C}
\exp
\left(
g_{m_1,d}
\left(
\boldsymbol{z}_{m_1}
\right)
\right)
},
\end{equation}
and
\begin{equation}
\footnotesize
p_{F}
\left(
c
\mid
\hat{\boldsymbol{z}}_{F\mid m_1,k}
\right)
=
\frac{
\exp
\left(
g_{F,c}
\left(
\hat{\boldsymbol{z}}_{F\mid m_1,k}
\right)
\right)
}{
\sum_{d=1}^{C}
\exp
\left(
g_{F,d}
\left(
\hat{\boldsymbol{z}}_{F\mid m_1,k}
\right)
\right)
}.
\end{equation}
Therefore, the approximated recovery prediction can be interpreted as a mixture of category-conditioned fusion predictions, where each component is weighted by the posterior probability of its corresponding category assumption:
\begin{equation}
\footnotesize
p_{m_1\rightarrow F}(c|\boldsymbol z_{m_1})
\approx
\sum_{k=1}^{C}
\frac{
\exp
\left(
g_{m_1,k}(\boldsymbol z_{m_1})
\right)
}{
\sum_{d=1}^{C}
\exp
\left(
g_{m_1,d}(\boldsymbol z_{m_1})
\right)
}
\,
p_F(c|\hat{\boldsymbol z}_{F|m_1,k}).
\end{equation}
This formulation indicates that the posterior obtained from the observed modality determines the contribution of each category-conditioned recovered fusion representation. Since the log-posteriors preserve the additive category evidence before normalization, we further perform the aggregation in the log-posterior space rather than directly averaging normalized probabilities. Specifically, the recovery logits are defined as
\begin{equation}
\footnotesize
\begin{aligned}
\boldsymbol g_{m_1\rightarrow F}
(\boldsymbol z_{m_1})
&=
\sum_{k=1}^{C}
p_{m_1}(k|\boldsymbol z_{m_1})
\boldsymbol g_F
(
\hat{\boldsymbol z}_{F|m_1,k}
)\\
&=
\sum_{k=1}^{C}
\frac{
\exp
\left(
g_{m_1,k}(\boldsymbol z_{m_1})
\right)
}{
\sum_{d=1}^{C}
\exp
\left(
g_{m_1,d}(\boldsymbol z_{m_1})
\right)
}
\boldsymbol g_F
(
\hat{\boldsymbol z}_{F|m_1,k}
).
\end{aligned}
\end{equation}

This is the recovery term used in the missing-modality calibration described in the main text.
The case where modality $m_1$ is missing follows analogously by taking $m_2$ as the observable modality.

\section{Sensitivity Analysis of Adaptation Batch Size}
\label{appendix:batchsize}

\begin{figure}[t]
  \centering
  \includegraphics[width=0.9\linewidth]{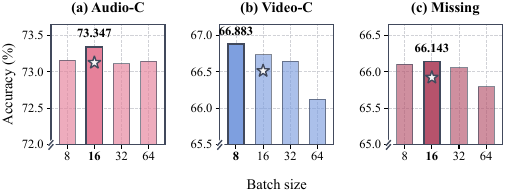}
  \vspace{-0.3cm}
  \caption{
Effect of batch size on Kinetics50. 
(a–c) Mean accuracy under audio corruption, video corruption, and modality-missing settings, respectively. 
Darker bars and bold values indicate the best performance in each setting, while stars mark the adopted batch size of 16.
}
  \label{fig:batchsize}
\end{figure}

We investigate the sensitivity of AdaPGC to the adaptation batch size. 
As shown in Figure~\ref{fig:batchsize}, increasing the batch size does not consistently improve performance. 
A batch size of 16 achieves the highest average accuracy under audio corruption and modality missing, while a batch size of 8 performs slightly better under video corruption, with a difference of 0.15 percentage points. 
Further increasing the batch size to 32 or 64 yields lower average accuracy across all three settings than using 16. 
These results support a batch size of 16 as a consistent choice across different distribution shifts, balancing performance under modality corruption and missingness. Therefore, we adopt a batch size of 16 in our experiments.

\end{document}